\documentclass[final,5p,times,twocolumn]{elsarticle}

\usepackage{listings}
\usepackage{graphicx}
\usepackage{booktabs} 
\usepackage{array,multirow}
\usepackage{url} 
\usepackage{natbib}
\usepackage{amssymb}
\usepackage{lscape}
\usepackage{mathrsfs}
\usepackage{subfig}
\usepackage{amsmath}
\usepackage{tikz}
\usepackage{float}
\usepackage{rotating}
\usepackage{ulem}
\usepackage{colortbl} 

\usepackage{hyperref}
\usepackage{tikz}
\usetikzlibrary{arrows,shapes,positioning,shadows,trees}
\usepackage{todonotes}
\usepackage{soul}
\usepackage{xcolor} 

\usepackage[ruled,linesnumbered]{algorithm2e}  

\usetikzlibrary{fit,calc}

\newcommand*{\tikzmk}[1]{\tikz[remember picture,overlay,] \node (#1) {};\ignorespaces}

\newcommand{\boxit}[1]{\tikz[remember picture,overlay]{\node[yshift=3pt,fill=#1,opacity=.25,fit={(A)($(B)+(.80\linewidth,.8\baselineskip)$)}] {};}\ignorespaces}

\newcommand{\boxitt}[1]{\tikz[remember picture,overlay]{\node[yshift=3pt,fill=#1,opacity=.25,fit={(A)($(B)+(.838\linewidth,.8\baselineskip)$)}] {};}\ignorespaces}

\colorlet{pink}{red!40}
\colorlet{blue}{cyan!60}
\colorlet{yellow}{yellow!60}

\tikzset{
  basic/.style  = {draw, text width=4cm, drop shadow, font=\sffamily, rectangle},
  root/.style   = {basic, rounded corners=2pt, thin, align=center,fill=green!60},
  level 2/.style = {basic, rounded corners=6pt, thin,align=center, fill=green!30,text width=8em},
  level 3/.style = {basic, thin, align=left, fill=pink!60, text width=6.5em},
  level 3a/.style = {basic, thin, align=left, fill=pink!30, text width=6.5em},
  level 4/.style = {basic, thin, align=left, fill=gray!40, text width=6.5em},
  level 5/.style = {basic, thin, align=center, fill=gray!5,text width=6.5em}
}

\makeatletter
\providecommand{\doi}[1]{%
	\begingroup
	\let\bibinfo\@secondoftwo
	\urlstyle{rm}%
	\href{http://dx.doi.org/#1}{%
		doi:\discretionary{}{}{}%
		\nolinkurl{#1}%
	}%
	\endgroup
}
\makeatother

\usepackage{amssymb}

\journal{Future Generation Computer Systems} 

\begin{document}

\begin{frontmatter}



\title{Distributed Genetic Algorithm for Service Placement in Fog Computing Leveraging Infrastructure Nodes for Optimization}


\author{Carlos Guerrero\corref{mycorrespondingauthor}}
\ead{carlos.guerrero@uib.es}
\cortext[mycorrespondingauthor]{Corresponding author. Email: carlos.guerrero@uib.es.  Telf: +34 971 17 29 65}

\author{Isaac Lera\corref{}}
\ead{isaac.lera@uib.es}

\author{Carlos Juiz\corref{}}
\ead{cjuiz@uib.es}

\address{Crta. Valldemossa km 7.5, Palma, E07121, SPAIN}

\address[mymainaddress]{Computer Science Department, University of Balearic Islands}

\begin{abstract}
The increasing complexity of fog computing environments calls for efficient resource optimization techniques. In this paper, we propose and evaluate three distributed designs of a genetic algorithm (GA) for resource optimization in fog computing, within an increasing degree of distribution. The designs leverage the execution of the GA in the fog devices themselves by dealing with the specific features of this domain: constrained resources and widely geographical distribution of the devices. For their evaluation, we implemented a benchmark case using the NSGA-II for the specific problem of optimizing the fog service placement, according to the guidelines of our three distributed designs. These three experimental scenarios were compared with a control case, a traditional centralized version of this GA algorithm, considering solution quality and network overhead. The results show that the design with the lowest distribution degree, which keeps centralized storage of the objective space, achieves comparable solution quality to the traditional approach but incurs a higher network load. The second design, which completely distributes the population between the workers, reduces network overhead but exhibits lower solution diversity while keeping enough good results in terms of optimization objective minimization. Finally, the proposal with a distributed population and that only interchanges solution between the workers' neighbors achieves the lowest network load but with compromised solution quality.

\end{abstract}


\end{frontmatter}



\section{Introduction}

Distributed computing solves complex problems by decomposing them into smaller, more manageable tasks that can be executed in parallel on multiple computers, referred to as workers. These workers collaborate to carry out the assigned tasks under the supervision of a coordinator, which oversees the overall execution and facilitates the required coordination among the workers.

Fog computing facilitates the seamless integration of distributed computing~\cite{SRIRAMA2021439} by leveraging fog devices as workers. These devices enable the deployment of distributed computational tasks in a flexible and dynamic manner, bringing the execution closer to the end-users. Meanwhile, the cloud can be a good candidate to assume the role of the coordinator, harnessing its virtually limitless computing resources, high availability, and fault tolerance to ensure the uninterrupted operation of the system, even in the face of failures or disruptions.

Recent literature has demonstrated the effectiveness of genetic algorithms (GAs) in optimizing resource allocation within fog infrastructure~\cite{GUERRERO2022101094}. However, GAs often suffer from longer execution times, which pose a challenge in highly dynamic environments like fog computing. To mitigate this drawback, distributed execution of optimization algorithms has been proposed. While some studies have explored distributed and parallel approaches for executing GAs in fog domains~\cite{10.1007/978-3-030-00374-6_24}, to the best of our knowledge, no previous research has investigated the utilization of fog infrastructure resources for its own optimization. 

Our proposal aims to utilize the inherent devices within the fog infrastructure to execute the fog resource optimization algorithms, facilitating their distributed execution. We name this as \textit{in-situ} deployment, which entails executing fog applications and operations management algorithms directly within the fog devices. This stands in contrast to dedicated or external computational nodes responsible for executing these optimization algorithms.

The deployment of a distributed algorithm in a fog domain necessitates careful consideration of its unique characteristics, which inherently differ from cloud computing. These distinctions arise from the fog infrastructure's larger scale, wider geographical dispersion of device locations, more constrained device resources, and heterogeneous node interconnections~\cite{GUERRERO2022101094,moysiadis2018}. Moreover, the intensive data exchange among computing instances within the distributed algorithm can lead to increased network traffic within the infrastructure. Our proposal will consider finding a trade-off to alleviate this network overhead without damaging the quality of the optimized solutions.

This paper investigates three design alternatives for the optimization algorithm, each exhibiting an increasing degree of distribution across the fog infrastructure. The extent of distribution determines how the task execution and solution space storage are assigned among the coordinator and/or the workers. Specifically, we have developed a fog-aware distributed version of NSGA-II (Elitist Non-Dominated Sorting Genetic Algorithm) \cite{deb2002fast}. NSGA-II has shown excellent performance in solving problems with two and three objectives, surpassing many other algorithms in terms of solution quality and convergence speed \cite{zitzler2001spea2,von2014survey}. The three proposed designs emphasize both solution quality and network overhead by striking a balance between the degree of distribution and optimization performance.

The three design alternatives for NSGA-II are implemented using Message Queuing Telemetry Transport (MQTT), a lightweight communication protocol suitable for fog computing domains. Between all the common resource optimization problems in fog domains, we select the problem of optimizing the application placement in fog devices (FAPP) as a benchmark for evaluating and comparing the three alternatives. To assess the performance of our proposals, we execute them to optimize the same infrastructure and we measure both network usage and solution quality achieved through the distributed execution of NSGA-II. Furthermore, we compare the results obtained from our three proposals with those obtained from a centralized version of NSGA-II.

The paper's main contributions can be summarized as follows:
\begin{itemize}
\item We propose an \textit{in-situ} deployment approach, utilizing the fog infrastructure's own devices for executing the optimization algorithm, thereby optimizing resource allocation within the fog architecture.
\item We present three distributed models for executing NSGA-II, responsible for infrastructure optimization, with varying degrees of distribution and reduced impact on network overhead.
\item We introduce a lightweight deployment of the distributed version of NSGA-II through a communication based on the MQTT protocol.
\item We evaluate the trade-off between optimization quality and network overhead of the three proposed distributed models and compare them with a control case, involving the traditional centralized execution of NSGA-II.
\end{itemize}

The rest of the paper is organized as follows: Section~\ref{sect_relatedwork} summarizes the current state of the art based on three research domains in the field of fog computing (resource management, distributed GAs, and \textit{in-situ} deployment); Section~\ref{sect_distributedproposals} explains the three distributed designs, with an incremental distribution degree, that we propose and evaluate in this paper and compares them with a traditional centralized execution of a GA; Section~\ref{sect_implementationdesign} presents the implementation details of the distributed proposals, mainly focusing on lightweight communication between the fog devices; Section~\ref{sect_evaluation} describes the experiments performed for the evaluation of the distributed GA proposals and analyzes the results of their execution in terms of solution quality and network load; and finally, Section~\ref{sect_conclusions} presents the conclusions and future work that arise from the findings of this study.

\section{Related work}
\label{sect_relatedwork}

\begin{figure}[h!]
	\centering
\scalebox{.6}{
	\begin{tikzpicture}[scale=0.750]
	\begin{scope}
    \draw[fill={rgb:black,1;white,10}] (5,2.2) circle (8cm);
	\draw[fill=white] (5,4.2) circle (4cm);
	\draw[fill=white] (3,0.7) circle (4cm);
	\draw[fill=white] (7,0.7) circle (4cm);
	\begin{scope}
	\clip (7,0.7) circle (4cm);
	\fill[blue!40!white] (3,0.7) circle (4cm);
	\end{scope}
	\begin{scope}
	\clip (7,0.7) circle (4cm);
	\fill[red!40!white] (5,4.2) circle (4cm);
	\end{scope}
	\begin{scope}
	\clip (3,0.7) circle (4cm);
	\fill[green!40!white] (5,4.2) circle (4cm);
	\end{scope}
	\begin{scope}
	\clip (3,0.7) circle (4cm);
	\clip (7,0.7) circle (4cm);
	\fill[yellow!40!white] (5,4.2) circle (4cm);
	\end{scope}
	
	\end{scope}
	
	\begin{scope}
    \node[text width=8cm, align=left] at (3,4.4) {Fog\\ Computing};
	\node[text width=4cm, align=center] at (5,6.2) {Resource\\ Management};
	\node[text width=4cm, align=center] at (1,0.7) {Distributed\\ GAs };
	\node[text width=4cm, align=center] at (9,0.7)  {In-situ\\ deployment };
	\end{scope}

	\begin{scope}
	\node[text width=4cm, align=center] at (5,5) {\cite{ghobaei2020resource}~\cite{GUERRERO2022101094}~\cite{9194714}~\cite{10.1145/3326066}};
	\end{scope}

	\begin{scope}
	\node[text width=4cm, align=center] at (2.5,3.25) {\cite{7776569}~\cite{9c434213a47d45b486ebc9c416d132d0} \\ \cite{7867735}};
	\end{scope}

	\begin{scope}
	\node[text width=4cm, align=center] at (7.5,3.25) {\cite{7405527}~\cite{8651835}\\ \cite{FORTI2021605}};
	\end{scope}

	\begin{scope}
	\node[text width=4cm, align=center] at (1,-0.5) {\cite{10.1145/3400031}};
	\end{scope}

	\begin{scope}
	\node[text width=4cm, align=center] at (9,-1) {\cite{10.1145/3301443}~\cite{8444370}\\ \cite{BITTENCOURT2018134}};
	\end{scope}

	\begin{scope}
	\node[text width=4cm, align=center] at (5,-0.5) {\cite{10.1007/978-3-030-00374-6_24}};
	\end{scope}

	\begin{scope}
	\node[text width=4cm, align=center] at (5,2) {[This\\ work]};
	\end{scope}

	\end{tikzpicture}

}

	\caption{Research domains of the related work and the research gap covered with our proposal.}\label{fig_researchdomains}
\end{figure}
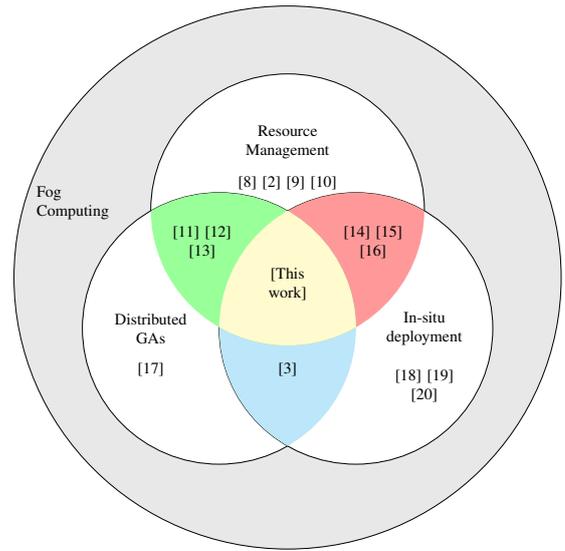

This paper tackles an optimization problem in the domain of fog computing that lies at the intersection of three primary research areas (\figurename{~\ref{fig_researchdomains}}): fog resource management, distributed GAs in fog domains, and the execution of operational optimization algorithms within the same infrastructure (fog devices) that requires to be optimized/managed (\textit{in-situ} deployment).

The research area of distributed and parallel GAs has garnered significant interest, resulting in a substantial body of literature. Notably, Harada and Alba's survey~\cite{10.1145/3400031} is a recent and comprehensive overview of this field. Their work examines well-established models and their recent implementations (2014-2020), while also introducing new open challenges and providing valuable insights to guide future research endeavors.

The implementation of distributed or parallel versions of GAs effectively addresses the primary limitation of these algorithms in terms of execution times, especially when dealing with large-scale optimization problems~\cite{cantu1998survey}. GAs are well-suited for parallelization and distribution due to their repetitive nature, which involves performing similar operations on multiple solutions or individuals within the population. The adoption of distributed and parallel GAs not only helps overcome scalability challenges but also facilitates the implementation of novel optimization models, techniques, and operators that can leverage parallel and heterogeneous platforms~\cite{alba2005parallel, talbi2013metaheuristics}. Additionally, distributed GAs offer enhanced compatibility with other search procedures compared to their sequential counterparts~\cite{alba1999survey}. It is worth noting that distributed GAs can exhibit improved search capabilities, even without the use of parallel hardware, demonstrating higher efficiency and efficacy compared to sequential GAs~\cite{alba1999survey}.

The research interest in fog resource management has also seen a significant rise. Ghobaei-Arani et al.\cite{ghobaei2020resource} conducted an analysis of papers in various areas such as application placement, resource scheduling, task offloading, load balancing, resource allocation, and resource provisioning. They compared these papers based on performance metrics, case studies, utilized techniques, and evaluation tools. Similarly, Guerrero et al.\cite{GUERRERO2022101094} surveyed papers that specifically employ GAs for optimizing resource management in fog environments. Martinez et al.~\cite{9194714} conducted a survey of papers focused on resource provisioning, allocation, and management, among other fields, with the goal of presenting a sequential organization of the necessary phases for implementing practical fog computing systems. Furthermore, Hong and Varghese~\cite{10.1145/3326066} reviewed relevant papers to identify and classify architectures, infrastructure, and underlying algorithms used for resource management in fog computing.

Existing surveys reflect the considerable research effort dedicated to exploring distributed GAs and fog resource management, covering their current state-of-the-art. However, the specific combination of distributed GAs in the context of resource management for fog computing necessitates a more in-depth analysis within the scope of our work, offering unique insights and contributions in this context.

In recent literature, several papers have explored the implementation of distributed genetic approaches, leveraging tools from the distributed big data ecosystem, to address resource optimization in fog computing domains. Mennes et al. utilized MongoDB~\cite{7776569} to implement a genetic algorithm for optimizing the placement of applications into the fog clusters. Yang et al.\cite{9c434213a47d45b486ebc9c416d132d0}, similarly to Wen et al.\cite{7867735}, focused on optimizing fog service orchestration using a distributed version of a genetic algorithm implemented with Spark. 

However, it is important to note that these previous solutions relied on external and dedicated resources for the execution of the optimization process, failing to leverage the distributed nature of the fog architecture. Furthermore, these approaches utilized resource-consuming platforms, and they were not suitable for \textit{in-situ} optimization due to the requirement of lightweight execution in fog devices.

In the context of \textit{in-situ} deployment of distributed GAs within a fog infrastructure, Morell and Alba~\cite{10.1007/978-3-030-00374-6_24} designed a GA version specifically tailored for execution on portable devices. Their study primarily focused on evaluating resource usage and execution time on highly resource-constrained devices such as smartphones, routers, and wearables. They concluded that while generic optimization benchmarks can provide an initial impression of performance on edge devices, single-board devices like the Raspberry Pi demonstrated better overall performance.

It is worth noting that our proposal differs from the work of Morell and Alba in several aspects. Firstly, we do not limit the execution of optimization algorithms to edge devices. Instead, our approach allows any intermediate fog devices, including more powerful ones, to participate in the optimization execution. This broader scope enables the utilization of a wider range of resources within the fog infrastructure. Additionally, our focus is specifically on optimizing the fog infrastructure itself, rather than addressing generic optimization problems. By concentrating on the unique characteristics and requirements of fog computing, we aim to develop tailored solutions that leverage the inherent capabilities of the fog architecture.

From the perspective of \textit{in-situ} deployment, there is currently no comprehensive survey exclusively dedicated to the deployment of applications in fog nodes, regardless of whether it focuses on user-specific applications or infrastructure operations. However, some more general surveys have included sections that touch upon this specific topic~\cite{10.1145/3301443,8444370,BITTENCOURT2018134}. Nevertheless,  several research efforts have focused on the distributed and self-organized management of tasks within fog architectures, leveraging the capabilities of the fog infrastructure. These works cover various types of tasks, such as resource scheduling~\cite{7405527}, service provisioning~\cite{8651835}, and monitoring~\cite{FORTI2021605}. However, to the best of our knowledge, none of the papers in this set have implemented distributed GAs to be executed \textit{in-situ} within the fog infrastructure. Therefore, there is still a research gap to be filled in terms of exploring the potential of distributed GAs for \textit{in-situ} optimization within the fog architecture.

In summary, our paper stands out from the related literature presented in \figurename{~\ref{fig_researchdomains}} by introducing a novel approach for optimizing fog infrastructures. It is the first proposal that implements a distributed GA executed \textit{in-situ}, meaning that the optimization process takes place within the same fog devices that execute the fog applications. Our research specifically addresses the specific characteristics of our problem domain, including the geographic distribution of fog devices, their connectivity through heterogeneous communication networks, and the challenge of achieving high-quality results while operating within the constraints of resource-constrained devices with limited storage capacity for solutions. By considering these specific features, our work aims to provide valuable insights into the optimization of fog infrastructures and pave the way for future advancements in this field.

\section{Distributed proposals for genetic optimization based on NSGA-II}
\label{sect_distributedproposals}

NSGA-II (Elitist Non-Dominated Sorting Genetic Algorithm) \cite{deb2002fast} is a highly regarded genetic algorithm commonly employed for addressing multi-objective optimization problems by leveraging the concept of Pareto dominance. The selection of NSGA-II is motivated by its capability to handle objective functions that are nonlinear and non-convex, its ability to identify multiple Pareto-optimal solutions, and its capacity to incorporate domain-specific knowledge. Furthermore, NSGA-II exhibits exceptional performance in scenarios involving two and three objectives, surpassing many alternative algorithms in terms of solution quality and convergence speed \cite{zitzler2001spea2,von2014survey}.

In a genetic algorithm (GA), the process of selecting solutions involves determining their order. In single-objective problems, this ordering is typically based on a scalar value associated with the objective. However, in multi-objective optimization, where there are multiple objectives to consider, the Pareto dominance concept is used to establish the ordering of non-scalar solutions~\footnote{A solution $s_1$ non-dominates solution $s_2$ if $s_2$ is not better for any objective and $s_1$ is better for at least one objective on the vector of objective values they possess.}. 

In the context of NSGA-II, the algorithm employs the Pareto dominance concept to recursively classify solutions into fronts. A front consists of solutions that are not dominated by any other solution and are not already included in previous fronts. Solutions in earlier fronts are considered to have a higher level of goodness than solutions in subsequent fronts. As a result, solutions in earlier fronts should be selected with a higher probability during the selection process. By using this approach, NSGA-II is able to identify a diverse set of non-dominated solutions along the Pareto front, providing a comprehensive view of the trade-offs between different objectives.

\begin{algorithm}[t!]
 \caption{Multi-objective genetic optimization algorithm~\cite{deb2002fast}}
 \label{nsga2}
    $P_t \gets generateRandomPopulation(pop_{size})$\\
    $fitness_{off} \gets calculateFitness(P_{t}) $\\
    $fronts \gets calculateFronts(P_t,fitness)$\\
    $distances \gets calculateCrowding(P_t,fronts,fitness)$\\

    \For(){$i\ $in$\ 1..gen_{num}$}{\label{alg_foroffspring}  
        $P_{off} \gets \emptyset$ \\
        \For(){$j\ $in$\ 1..pop_{size}$}{
            \tikzmk{A}$father1 \gets binaryTournament(P_t,fronts,distances)$\label{alg_tournament1}\\
            $father2 \gets binaryTournament(P_t,fronts,distances)$\label{alg_tournament2}\\\tikzmk{B}\boxit{pink}
            \tikzmk{A}$child1,child2 \gets crossover(father1,father2)$\label{alg_crossovercode}\\
            \If {$random(0,1) < \rho_{mut}$}{\label{alg_mutationcode1}                              
                $mutation(child1)$,$mutation(child2)$\\
    	}\label{alg_mutationcode2}
     
            $fitness1 \gets calculateFitness(child1)$\label{alg_fitness1}\\
            $fitness2 \gets calculateFitness(child2)$\label{alg_fitness2}\\ \tikzmk{B}\boxit{blue}
            \tikzmk{A}$fitness_{off} \gets fitness_{off} \cup \{fitness1, fitness2\} $\label{alg_combinefitness}\\
            $P_{off} \gets P_{off} \cup \{child1, child2\} $\label{alg_combinepop}\\\tikzmk{B}\boxit{yellow}
        }
        \tikzmk{A}$P_{union} \gets P_{off} \cup P_{t}$  \label{alg_join}\\
        $fronts \gets calculateFronts(P_{union},fitness)$\label{alg_calculatefront} \\
        $distances \gets calculateCrowding(P_{union},fronts,fitness)$\label{alg_calculatedistance}\\
        $P_{sorted} \gets sortElements(P_{union},fronts,distances)$ \label{algo_elitism1}\\
        $P_{t} \gets P_{sorted}[1..pop_{size}]$  \label{alg_half} \label{algo_elitism2}\\\tikzmk{B}\boxitt{yellow} 
    }
    $Solution \gets fronts[1]$ \#the Pareto front\\
\end{algorithm}

Algorithm~\ref{nsga2} presents the typical structure of NSGA-II, which iteratively generates new offspring populations. NSGA-II generates new solutions by selecting the best solution from the current population and applying genetic operators such as crossover and mutation to these solutions. The resulting solutions are then classified based on non-domination sorting and crowding distance. In a distributed fog computing scenario, factors such as storage organization and solution retrieval play a role in the selection and sorting processes. The algorithm can be divided into the following main blocks:
\begin{itemize}
    \item Parent selection (red box in Algorithm~\ref{nsga2}): NSGA-II utilizes the binary tournament selection operator, where two solutions are randomly selected from the population, and the better one is chosen as the first parent. The second parent is selected in the same manner (lines \ref{alg_tournament1}--\ref{alg_tournament2} in Algorithm~\ref{nsga2}). In a distributed scenario, the organization of population storage influences this selection process and the retrieval of solution data by the workers.
    \item Children generation  (blue box in Algorithm~\ref{nsga2}): After selecting the two parents and retrieving their data by the workers, the solutions undergo mating using the crossover operator (line \ref{alg_crossovercode} in Algorithm~\ref{nsga2}). Additionally, random alterations are introduced through the mutation operator (lines \ref{alg_mutationcode1}--\ref{alg_mutationcode2} in Algorithm~\ref{nsga2}). Finally, the fitness values (objective values) for the two resulting children are calculated (lines \ref{alg_fitness1}--\ref{alg_fitness2} in Algorithm~\ref{nsga2}).
    \item Offspring analysis  (yellow boxes in Algorithm~\ref{nsga2}): The previous steps are repeated until a new offspring is obtained (line \ref{alg_foroffspring} in Algorithm~\ref{nsga2}), and these new solutions are combined with the current population (lines \ref{alg_combinefitness}--\ref{alg_combinepop} in Algorithm~\ref{nsga2}). The next generation's population is derived by organizing the solutions into fronts using the non-domination concept and ordering the solutions within each front based on the crowding distance (lines \ref{alg_calculatefront}--\ref{algo_elitism2} in Algorithm~\ref{nsga2}). In a distributed environment, the storage locations of the solutions influence this sorting process.
\end{itemize}

In a conventional centralized execution of NSGA-II, a single computer or node handles the execution of the three blocks and manages the storage of the solution space, which includes the decision and objective spaces. However, the literature has also proposed distributed designs for population-based algorithms~\cite{10.1145/3400031}.

Two commonly used distributed models are the global parallelization model and the parallel island model. In the global parallelization model, solutions are managed in a centralized population, while the evaluation of solutions is performed in multiple workers. On the other hand, the parallel island model involves managing solutions in separate islands or distributed sub-populations. Each island operates independently, executing its own instance of the GA. Periodic interchange of solutions between islands facilitates information exchange and exploration of different regions in the search space.

It is worth noting that these models have been extensively used in traditional datacenters, where nodes have (\textit{virtually}) unlimited resources and are connected via high-speed networks. 
In contrast, our proposal aims to leverage the resources of the fog infrastructure itself for optimization. Specifically, we have developed a distributed version of NSGA-II, where the fog devices act as workers and the cloud serves as the coordinator. Our approach offers three distributed design alternatives, each progressively increasing the degree of distribution. These design alternatives determine which tasks or steps of the GA execution are performed in the fog nodes (workers) or in the cloud (coordinator). 

\begin{table*}[t!] \footnotesize
	\caption{Tasks distribution for the considered scenarios.}
	\label{tab_distributionschema}
	\centering
	\begin{tabular}{lp{2cm}|p{2cm}|p{2cm}|p{2cm}|p{2cm}}
		\toprule
		&&\textbf{Traditional}&\textbf{Semi-}&\textbf{Fully-}&\textbf{Neighbor-}\\
  		&&\textbf{}&\textbf{distributed}&\textbf{distributed}&\textbf{aware}\\
		\midrule

      \multirow{2}{*}[-0.5em]{\rotatebox{90}{\parbox{0.7cm}{ Pop.\\ storage}}}
   & Decision space &  \cellcolor{yellow!25} coordinator & \cellcolor{blue!25}workers & \cellcolor{blue!25}workers & \cellcolor{blue!25}workers \\ \cmidrule{2-6}
   & Objective space & \cellcolor{yellow!25}coordinator & \cellcolor{yellow!25}coordinator & \cellcolor{blue!25}workers & \cellcolor{blue!25}workers \\\midrule
	&	 1st parent selection & \cellcolor{yellow!25}coordinator & \cellcolor{blue!25}locally in the workers & \cellcolor{blue!25}locally in the workers & \cellcolor{blue!25}locally in the workers\\ \midrule
	&	 2nd parent selection & \cellcolor{yellow!25}coordinator & \cellcolor{yellow!25}coordinator & \cellcolor{blue!25}in a random remote worker & \cellcolor{blue!25}in a random remote neighbor \\ \midrule
	&	 Children generation & \cellcolor{yellow!25}coordinator & \cellcolor{blue!25}workers & \cellcolor{blue!25}workers & \cellcolor{blue!25}workers \\ \midrule
	&	 Offspring analysis & \cellcolor{yellow!25}only in the coordinator &  \cellcolor{yellow!25}complete in the coordinator and partially in the workers & \cellcolor{blue!25}partially in the workers & \cellcolor{blue!25}partially in the workers \\

		\bottomrule
	\end{tabular}
\end{table*}

Table~\ref{tab_distributionschema} provides a comprehensive overview of our proposed designs and compares them with the traditional centralized execution. The table outlines the task blocks performed by the coordinator and the workers, as well as the storage approach for both components of the solution space: the decision space $\mathbb{X}$ (the set of solutions generated by the optimization algorithm that are composed by the specific values to the decision variables) and the objective space $\mathbb{Z}$ (the objective values associated with each solution and obtained by the evaluation of the defined optimization objectives).

Our three distributed designs for executing optimization algorithms in a fog infrastructure are designed in a progressive manner, gradually transferring execution blocks and population storage from the centralized coordinator to the distributed workers. These approaches takes into account the unique characteristics of fog computing, which distinguish it from other distributed architectures. These characteristics include a higher scale level, wider geographical dispersion of device locations, limited resources in the devices, and heterogeneity in node interconnection~\cite{GUERRERO2022101094,moysiadis2018}. These specific features of fog domains play a crucial role in shaping our design decisions, which differ from previous distributed proposals for GA execution~\cite{10.1145/3400031,1041554,Sudholt2015}.

Our first design proposal, namely as the \textit{Semi-distributed} model, represents an intermediate approach between the global and island models. It aims to strike a balance between minimizing network usage and enabling the global evaluation of the solution space. In this design, mating, mutation, and solution evaluation are performed by the workers, similar to the island model.

However, in order to alleviate network workload, we adopt a partially centralized storage approach. Unlike the global model where both the decision space and objective space are stored centrally, in our \textit{Semi-distributed} model, only the objective space is stored in the coordinator. On the other hand, the solution space remains distributed among the workers. 

A centralized storage of the objective space in our \textit{Semi-distributed} design allows us to maintain a panmictic structure of the population, where each individual has all the other individuals as potential parents~\cite{Sudholt2015}. This centralized storage enables the calculation of a global Pareto front, which is beneficial for the continuous interchange of solutions.

In island models, the selection of mated solutions is typically limited to within each sub-population. However, it has been observed that continuous interactions between sub-populations are crucial for the performance of distributed GAs, as some designs may underperform compared to their non-distributed counterparts, particularly when dealing with small population sizes~\cite{alba2000influence}.

The deployment of our proposed GA designs should result in sub-populations with a highly limited number of solutions due to the resource constraints of fog devices. Therefore, in our first design, we employ a random selection of individuals from any sub-population for each mating operation. By maintaining a global Pareto front through the centralized storage of the objective space, we further enhance the probability of selecting the best solutions from across the entire population.

In summary, the \textit{Semi-distributed} proposal (Table~\ref{tab_distributionschema}) stores the decision space in a distributed manner across the workers, while a centralized copy of the objective space is maintained in the coordinator. This configuration allows for the calculation of a global Pareto front in the coordinator, as well as partial Pareto fronts within each sub-population in the workers. 

During the mating operation, one parent is selected from the global Pareto front, while the other parent is selected locally within the worker using the partial Pareto front. The selected parent from the global Pareto front is transmitted from the storing worker to the worker responsible for selecting the local parent. The mating operation, including crossover, mutation, and fitness evaluation, is then performed in this second worker.

Once the fitness is evaluated, the objective values are transmitted back to the coordinator to update the objective space and the global Pareto front. This exchange of information between the workers and the coordinator ensures that the objective space and the global Pareto front reflect the most up-to-date information from the distributed solution space.

In the second design proposal, known as the \textit{Fully-distributed} approach, we aim to further increase the level of distribution and reduce the reliance on the global Pareto front. In this design, the solution space is stored fully distributed among the workers, eliminating the need for a centralized storage of the solution space. As a result, each worker maintains its own partial Pareto front based on its locally stored solutions. This approach reduces the network usage because the solution objectives are not transmitted to the central population.

This second proposal, the \textit{Fully-distributed}, still differs from the traditional island model in terms of the continuous interchange of solutions between workers for each mating operation. In this design, the worker performing the mating operation selects a solution randomly from its own partially ordered subset of solutions (Table~\ref{tab_distributionschema}). For the second parent, the worker uniformly and randomly chooses a second worker. This second worker randomly selects one solution from its ordered set of solutions and transmits the second father to the first worker to initiate the mating operation. After the crossover, mutation, and fitness calculation are performed in the first worker, this updates its subset of solutions, including the recalculation of fronts. This process is repeated to enable ongoing mating operations within the \textit{Fully-distributed} environment.

This solution selection design in the \textit{Fully-distributed} approach retains the panmictic nature of the population, but it introduces a potential trade-off by maintaining only partial orders of individuals within the entire population. Consequently, there might be a reduction in the probability of selecting the best solutions compared to the \textit{Semi-distributed} approach. Our study aims to quantify this potential loss in solution quality and compare it with the benefits, such as reduced network usage, that arise from avoiding an up-to-date global and centralized storage of the objective space as proposed in our first design.

However, the \textit{Fully-distributed} proposal still faces challenges in terms of interchanging solutions between devices in the fog domain. Fog infrastructures are characterized by their extensive geographical distribution, and in some cases, the network distance between pairs of devices can be substantial. Furthermore, considering the heterogeneity and limited speed of network connections, the transmission of solutions between workers can significantly impact network performance. To address this issue, our third proposal, \textit{Neighbor-aware}, focuses on limiting the interchange of solutions to neighboring nodes, thereby reducing transmission overhead and network time. Various alternatives exist for defining the set of neighboring nodes in this proposal. For instance, neighboring nodes could be defined as devices directly connected to the worker via a single network edge. Alternatively, a threshold based on network distance or the number of hops between the worker and candidate neighbors could be used to determine the set of neighboring nodes.

It is anticipated that the quality of the final optimized solution set will gradually decrease from the first proposal to the last one. Conversely, the theoretical benefits in terms of reducing network overhead are expected to increase as the distribution degree increases across the three proposals. To provide the system administrator with decision criteria for choosing among the three alternatives, a thorough experimental evaluation is necessary. This evaluation will measure the benefits and drawbacks of each proposal, allowing the administrator to assess the balance between optimization quality and network usage based on their specific interests and requirements.

\section{Fog-based implementation of distributed designs for NSGA-II}
\label{sect_implementationdesign}

The implementation of the framework for executing the GA must consider the characteristics of fog domains, including limited resources and geographical distribution of nodes. Consequently, it is crucial to minimize resource consumption and network communication protocol overhead in the GA instances.

For communication between the GA instances, we utilize the MQTT (Message Queuing Telemetry Transport) protocol. MQTT is a lightweight message queue-based protocol specifically designed for domains with remote locations that have constrained device resources and/or limited network bandwidth, such as Internet of Things (IoT) or fog infrastructures~\cite{8969615}. To facilitate this communication, we have deployed Eclipse Mosquitto, an open-source implementation of the MQTT protocol.

MQTT is based on a publish-subscribe model, where clients connect to a broker and communicate through topics. In our implementation, the coordinator and workers of the GA instances communicate with each other through a message queue maintained by the MQTT broker. Topics serve as channels or subjects to which clients can receive (be subscribed to) or send (publish) messages. Clients can publish messages to specific topics, and the broker then distributes those messages to all subscribed clients interested in that topic. 

Implementing a distributed GA using MQTT involves identifying the data that needs to be shared between the GA instances (coordinator and workers), designing the sequence of messages to share this data, and defining MQTT topics. The data interchange between the GA instances differs in our first proposal compared to the other two. On the contrary, the two latter proposals only differs in the set of devices that interchange solutions, but their communication patterns are the same. Therefore, the design choices will be explained independently in a fist subsection for the \textit{Semi-distributed} model and for the other two proposals in the second subsections.

\subsection{Message communication for the \textit{Semi-distributed} GA proposal}

\begin{figure*} [t!]
\centering
\includegraphics[trim=0 0 0 0,clip,width=0.75\textwidth]{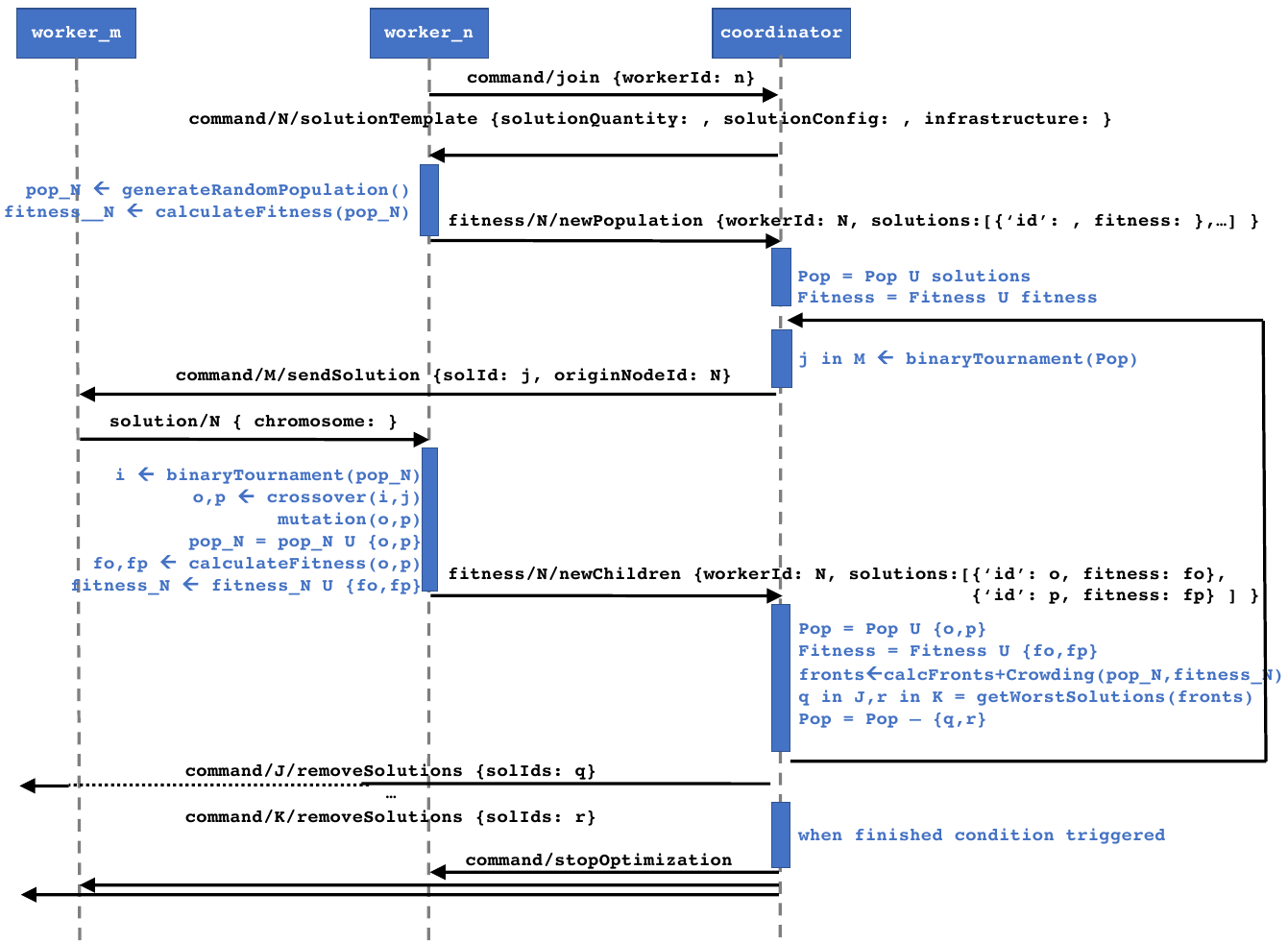}
\caption{Message diagram for the communication in the \textit{Semi-distributed} GA proposal. The optimization tasks executed by workers and the coordinator are labeled in blue color. The messages are labelled in black color with the format ''\textit{topic \{payload\}}''.}
\label{fig_semidistributedmessages}
\end{figure*}

Figure~\ref{fig_semidistributedmessages} illustrates the sequence of messages required for coordinating the GA instances and sharing the population data (decision and objective spaces) between them.

The process begins with an initialization period. During this phase, the workers notify the coordinator of their intention to join the optimization execution by publishing a message with the topic \verb|command/join|. Upon receiving this joining request, the coordinator publishes the initialization data using the topic \verb|command/workerID/solutionTemplate|. Subsequently, the workers generate the random initial population and publish the objective space of these initial solutions. The coordinator receives these objective spaces, published by the workers with the topic \verb|fitness/workerId/newPopulation|, and incorporates them into the global centralized objective space.

Once the initialization is completed, each worker enters an infinite loop until the coordinator publishes the end of the simulation using the topic \verb|command/stopOptimization|. During each iteration in the workers, which corresponds to a mating operation, two parents are selected to create two new children, and the population space is updated accordingly. This approach follows a steady-state reproduction model~\cite{SYSWERDA199194}, where the population is updated after the creation of each new solution. In contrast, the generational model updates the population once a complete new generation is obtained.

For each iteration, the first parent is selected from the global centralized population, while the second parent is chosen from the worker's local sub-population. The two parents are combined using the crossover operator and, if necessary, subjected to mutation to generate two new children. The fitness values of the children are then calculated. Finally, the local sub-population is updated to include the decision and objective spaces of the new children, while the centralized global population is updated only with the decision space of these children.

To update the global decision space, each iteration concludes by publishing a message to notify the coordinator about the creation of two new individuals using the topic \verb|fitness/workerId/newChildren|. This message also serves as an indication that the sender worker (represented as \textit{worker\_n} in Figure~\ref{fig_semidistributedmessages}) has completed the previous operation and is now idle, awaiting the start of a new mating operation. Upon receiving the message, the coordinator updates the decision space and triggers the start of a new mating operation.

The coordinator selects the first parent from the global centralized solution space and publishes a message to the worker responsible for storing the decision space of the selected solution (\textit{worker\_m} in Figure~\ref{fig_semidistributedmessages}). This message requests \textit{worker\_m} to send the selected solution to the idle worker (\textit{worker\_n}). When \textit{worker\_m} receives the message to send the solution (\verb|command/workerId/sendSolution|), it includes the chromosome of the solution in a message (\verb|solution/workerId|) and publishes it to the waiting worker, \textit{worker\_n}. Once the idle worker receives the chromosome of the first parent, it proceeds with a new mating operation.

The update of the global solution space may require the removal of certain solutions from the population. Throughout the optimization process, the population size remains constant. If either of the two children generated in a mating operation is superior to any of the current solutions in the population, the worst solutions need to be removed. These inferior solutions are eliminated from both the global objective space in the coordinator and their corresponding sub-populations in the workers. As a result, the coordinator notifies the workers to remove these solutions using the topic \verb|command/workerId/removeSolutions|.

Table~\ref{tab_semidistributedtopicdesign} provides a summary of the topics we have defined for the implementation of the \textit{Semi-distributed} GA. It includes a brief description of each topic, the MQTT client that publishes the message, and the MQTT clients that are subscribed to the specific topic and receive the message.

\begin{table*}[t!] \footnotesize
	\caption{Topic design for the communication in the \textit{Semi-distributed} GA proposal.}
	\label{tab_semidistributedtopicdesign}
	\centering
	\begin{tabular}{p{2,5cm}|l|l|p{10cm}}
		\toprule
		\textbf{Topic}&\textbf{Publisher}&\textbf{Subscribers}&\textbf{Description}\\
		\midrule

command/ nodeId/ solutionTemplate&coordinator&workers & Answer for a joining request. Attributes: the size of the population in the worker, the shape of the solutions' chromosome and the infrastructure data. \\\midrule
command/ nodeId/ removeSolutions&coordinator&workers& Notification of the solutions removed in global objective space. Attributes: identifiers of the solutions to remove in the workers. \\\midrule
command/ nodeId/ sendSolution&coordinator&workers& Request for the decision space data of a solution. Attributes: identifier of the solution and identifier of the node which needs the solution.\\\midrule
solution/ nodeId&workers&workers& Answer to a solution request. Attributes: the chromosome of the solution.\\\midrule
command/ stopOptimization&coordinator&workers& Notification to finish the optimization execution. No attributes.\\\midrule
command/ join&workers&coordinator& Request to join the optimization execution. Attributes: the identifier of the worker to join.\\\midrule
fitness/ nodeId/ newPopulation&workers&coordinator& Objective space data of the initial random population of a worker. Attributes: identifier of the worker, identifiers of the solutions and fitness values of the solutions.\\\midrule
fitness/ nodeId/ newChildren&workers&coordinator&Objective space data of the two children of a mating operation. Attributes: identifier of the worker, identifiers of the solutions and fitness values of the solutions.\\

		\bottomrule
	\end{tabular}
\end{table*}

\subsection{Message communication for the \textit{Fully-distributed} and \textit{Neighbor-aware} GA proposals}

The differences in implementation between the \textit{Fully-distributed} and \textit{Neighbor-aware} proposals are primarily related to the selection of parents, with the latter being limited to neighboring nodes of a worker. However, these differences do not impact the sequence of messages or the topic designs for communication between workers. Therefore, this section will detail the message sequence for both proposals without any variations.

\begin{figure*} [t!]
\centering
\includegraphics[trim=0 0 0 0,clip,width=0.75\textwidth]{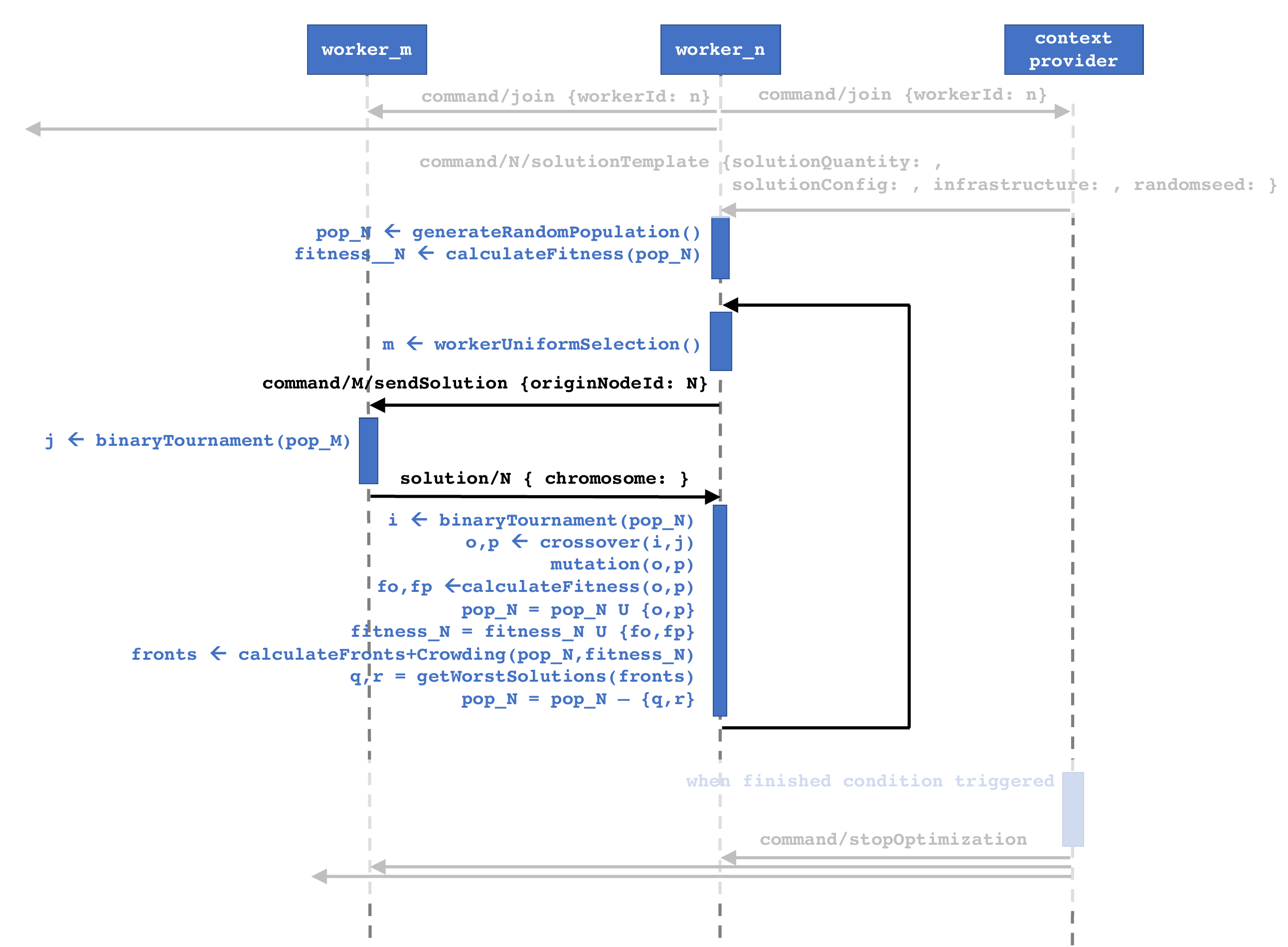}
\caption{Message diagram for the communication in the \textit{Fully-distributed} and \textit{Neighbor-aware} GA proposals. The optimization tasks executed by workers are labeled in blue color. The messages are labeled in black color with the format ''\textit{topic \{payload\}}''. The optional designs for the initialization and finish phases are blurred.}
\label{fig_fullydistributedmessages}
\end{figure*}

The communication in both the \textit{Fully-distributed} and \textit{Neighbor-aware} proposals is significantly simpler compared to the \textit{Semi-distributed} proposal (Figure~\ref{fig_fullydistributedmessages}). This is due to the absence of a centralized global solution space. As a result, the communication and data exchange are limited to the direct interchange of solutions between the workers responsible for storing the pair of parents to be mated, without involving an intermediate coordinator.

In the \textit{Fully-distributed} and \textit{Neighbor-aware} proposals, each worker independently initiates the mating operations. To start a new mating operation, a worker (e.g., \textit{worker\_n} in Figure~\ref{fig_fullydistributedmessages}) randomly selects another worker and sends a message with the topic \verb|command/workerId/sendSolution| to request the first parent solution. The selected worker (e.g., \textit{worker\_m} in Figure~\ref{fig_fullydistributedmessages}) receives the message and randomly selects a solution using its partially sorted sub-population. It then publishes the chromosome of the solution for the requesting worker (\textit{worker\_n}) using a message with the topic \verb|solution/workerId|. Upon receiving the solution's chromosome, \textit{worker\_n} begins the mating operation: selecting the second parent from its local sub-population, performing crossover, potentially mutating the children, calculating their fitness values, and updating only the local sub-population with the decision and objective spaces of the new children. To maintain a consistent population size throughout the optimization process, the worst solutions are removed from the local sub-population.

Note that although the iterative process of the mating operation is executed without the necessity of a coordinator, the initialization phase of the GA requires context data from the system. This context data, such as the resources of the infrastructure to optimize, the services that are executed, the number of users connected, etc., determines the template of the solutions of the GA. Consequently, this phase could be performed with some kind of context-awareness process in each worker that joins the execution of the optimization algorithm~\cite{DAS2023100049}. Alternatively, the context data could be provided by a central entity that stores these data, avoiding the context-aware discovering process. Any of those both implementation alternatives does not influence the design of our proposal.

Accordingly, we have supposed the existence of a lightweight central entity, namely context provider, which provides the initialization data to the workers that join the optimization. This is performed through messages with topics \verb|command/workerId/solutionTemplate| and \verb|command/join| (as depicted in the blurred zone in Figure~\ref{fig_fullydistributedmessages}). However, an alternative approach could involve a self-discovery process within each worker to obtain the necessary context data for the optimization execution. This self-discovery process would not affect the iterative mating operations or the interchange of the solution space. Additionally, we have assumed that the context provider also notifies the end of the optimization execution (\verb|command/stopOptimization|). Alternatively, a consensus mechanism could be used to determine the end of the optimization without impacting the design of our proposals.

Table~\ref{tab_fullydistributedtopicdesign} summarizes the topics defined for the \textit{Fully-distributed} and \textit{Neighbor-aware} GAs, including a brief description, the MQTT client responsible for publishing the message, and the MQTT clients subscribed to each specific topic to receive the message. The topics related to the optional initialization and finishing phases are also included in the table, smoothed with a gray background color.

\begin{table*}[t!] \footnotesize
	\caption{Topic design for the communication in the \textit{Fully-distributed} and \textit{Neighbor-aware} GA proposals.}
	\label{tab_fullydistributedtopicdesign}
	\centering
	\begin{tabular}{p{2,5cm}|l|l|p{10cm}}
		\toprule
		\textbf{Topic}&\textbf{Publisher}&\textbf{Subscribers}&\textbf{Description}\\
		\midrule

command/ nodeId/ sendSolution & worker & worker& Request for the decision space data of one solution in a random worker. Attributes: identifier of the node which needs the solution.\\\midrule

solution/ nodeId & worker & worker& Answer to a solution request. Attributes: the chromosome of the solution.\\\midrule

\cellcolor{gray!25}command/ stopOptimization  & \cellcolor{gray!25}context provider & \cellcolor{gray!25}worker&\cellcolor{gray!25} Notification to finish the optimization execution. No attributes.\\\midrule

\cellcolor{gray!25}command/ nodeId/ solutionTemplate & \cellcolor{gray!25}context provider & \cellcolor{gray!25}worker&\cellcolor{gray!25} Answer for a joining request. Attributes: the size of the population in the worker, the shape of the solutions' chromosome and the infrastructure data.\\\midrule

\cellcolor{gray!25}command/ join  & \cellcolor{gray!25}worker & \cellcolor{gray!25}context provider&\cellcolor{gray!25} Request to join the optimization execution. Attributes: the identifier of the worker to join.
\\

		\bottomrule
	\end{tabular}
\end{table*}

\section{Experimental evaluation}
\label{sect_evaluation}

The experimental evaluation aims to compare the performance of the three distributed genetic-based fog resource optimization proposals. To achieve this, a standard fog resource problem is chosen as a benchmark to assess the network usage and the quality of the optimization results across the three proposals. Additionally, the results obtained from the distributed proposals are compared against a control case, which involves a centralized monolithic implementation of the optimization algorithm. This allows for a comprehensive evaluation and analysis of the benefits and limitations of the distributed approaches in comparison to a centralized approach.

\subsection{Definition of the benchmark problem for the resource optimization}

The benchmark problem chosen for the comparison of the distributed GA proposals is the fog application placement problem (FAPP) in the resource management domain of fog computing. The placement of applications and services within a fog infrastructure is a challenging task, and it has received significant attention in the research literature~\cite{brogi2020place}. Many studies have explored various approaches to address this problem, and genetic-based solutions have been commonly considered for optimizing resource utilization in fog environments~\cite{GUERRERO2022101094}. 

FAPP can be formally defined as follows: Given a set of applications $A$, which need to be mapped onto a set of computational nodes $I$ belonging to a fog infrastructure, the goal is to find a mapping that satisfies a set of constraints $C$ while optimizing a set of objectives $O$.

In FAPP, each computational node or device $i \in I$ is characterized by its available resources for executing applications, denoted as $i_{resources}$. On the other hand, each application $a \in A$ is characterized by the resources it consumes when executed on a node, represented as $a_{consumption}$.

The solution mappings are considered to be many-to-many, meaning that an application can be placed on one or more nodes, and a node can host multiple applications. To represent these mappings, we define the decision variables as a matrix, where the elements $x_{a,i}$ represent the assignment of application $a \in A$ to node $i \in I$. In this matrix, if $x_{a,i}$ is equal to 1, it indicates that application $a$ is assigned to node $i$. Conversely, if $x_{a,i}$ is equal to 0, it means that application $a$ is not assigned to node $i$.

The set of users $U$ is connected to the fog devices acting as gateways in the fog infrastructure. The relationship between users and applications is a many-to-many relationship, meaning that users can request any of the applications present in the system and that applications can be requested by any user. To represent this requesting relationship between users and applications, we can use a matrix notation. Let's define a binary matrix $R$ of size $|A|\times|U|$, where $|A|$ represents the number of applications and $|U|$ represents the number of users. The elements $r_{a,u}$ in the matrix $R$ are equal to 1 if application $a \in A$ is requested by user $u \in U$, and $r_{a,u} = 0$ otherwise.

The objective functions in FAPP reflect the interests of the infrastructure operator and serve as indicators to be improved during the optimization process. GAs have been widely studied and validated as effective optimization techniques for this problem in the related literature~\cite{GUERRERO2019131}. 

In our study, our focus is on evaluating and comparing the distributed designs of the GA rather than exploring the specific benefits of genetic-based optimization. Therefore, we have selected two optimization objectives that are relatively easy to calculate and measure. Specifically, our genetic optimization for the benchmark problem aims to minimize the number of instances of applications and the network distance between users and fog devices executing their requested applications, which are opposite objectives.

Minimizing the number of instances of applications aims to reduce resource consumption in the fog infrastructure. By assigning multiple users to a single instance of an application, resources can be efficiently utilized, leading to cost savings and improved resource efficiency. Conversely, having a limited number of instances centralized in a few fog devices would decrease the number of instances but increase the network distance for some users.

Minimizing the network distance between users and applications aims to provide low-latency and responsive services to users. Replicating instances and placing them closer to users reduces the network delay and improves the overall user experience. This objective ensures that users can access the requested applications with minimal network latency, enhancing responsiveness and reducing communication overhead. Placing multiple instances of applications close to users would reduce network distance but increase the number of instances. 

The number of instances for a given application $a$, namely $instances(a)$, can be easily calculated by counting its number of instances:
\begin{gather}
instances(a) = \sum_{\forall i \in I}  x_{a,i}
\end{gather} We select the mean value of the number of instances for all the applications as the indicator for our first optimization objective, $o_1 \in O$: 
\begin{gather}
o_1 \Rightarrow \overline{instances} = \sum_{\forall a \in A} \frac{instances(a)}{|A|}
\end{gather}

We define the network distances between a given application $a$ and the set of users that request it, namely $distances(a)$, as the mean value of the distance between each of these users and the closest application instance, i.e. the minimum distance between the user and all the application instances. It is formerly defined as:
\begin{gather}
distances(a)=\frac{\sum_{\forall u \in U} r_{a,u} \times \min_{\forall x_{i,a}=1} d(x_{i,a},u)}{\sum_{\forall u \in U} r_{a,u}}
\end{gather}
where $d(x_{i,a},u)$ is the network latency between the device in which the user $u$ is connected and the fog device $i$ in which the application $a$ is instantiated.
Finally, the average of all the application distances is considered as an indicator of the second optimization objective, $o_2 \in O$:
\begin{gather}
o_2 \Rightarrow \overline{distances} = \sum_{\forall a \in A} \frac{distances(a)}{|A|}
\end{gather}

The mapping process is constrained by two conditions. The first one, $c_1 \in C$ is that one instance is at least required for each application: 
\begin{gather}
c_1 \Rightarrow \sum_{\forall i \in I}  x_{a,i} \geq 1, \forall a \in A
\end{gather}
Secondly, the sum of the resources demanded by the instances in a device should be smaller than the total resources available in that device, $c_2 \in C$:
\begin{gather}
c_2 \Rightarrow \sum_{\forall a \in A} x_{i,a} \times a_{consumption} \leq i_{resources}, \forall i \in I
\end{gather}

Given that, our optimization problem is summarized as finding the mapping between applications and fog devices:
\begin{gather}
x_{a,i}\ \forall a \in A \land \forall i \in I
\end{gather}
that minimizes the number of instances and the distances between users and applications:
\begin{gather}
    o_1 \Rightarrow  \overline{instances}\ \land\ o_2 \Rightarrow 
 \overline{distances} 
\end{gather}
considering that this mapping is constrained by having all least one application instance and by the resources available in the fog devices:
\begin{gather}
c_1 \land c_2
\end{gather}

\subsection{Experiment design}

For the experimentation, it is necessary to define the configuration of the GA, the features of the distributed execution and the infrastructure where the experiments will be conducted.

\begin{table}[t!]
	\caption{Genetic parameters for the experiments.}
	\label{tab_geneticparameters}
	\centering
	\begin{tabular}{p{3.5cm}|r}
		\toprule
		\textbf{Parameter}
		& \textbf{Value}\\
		\midrule
  \multicolumn{2}{l}{\textbf{Traditional genetic parameters}}  \\ \midrule
		 Population size &  200 \\ \midrule
		 Number of generations &  100 \\ \midrule
		 Mutation probability &  0.3 \\ \midrule
   \multicolumn{2}{l}{\textbf{Distributed execution parameters}}  \\ \midrule
		 Number of workers &  20 \\ \midrule
		 Sub-population size &  $\frac{\text{Population size}}{\text{Num. workers}}=10$ \\ \midrule
		 Neighborhood radius &  1 hop \\   
		\bottomrule
	\end{tabular}
\end{table}

The configuration parameters of a GA strongly influence the optimization process and they need to be carefully selected to ensure a proper balance between exploration and exploitation during the optimization process~\cite{birattari2009tuning}. The calibration of these parameters is typically performed during a pre-exploratory phase~\cite{10.1145/3377929.3398136}, in which a range of values is studied to determine the most suitable parameterization. In this exploratory phase, we studied and fixed the values of the total population size, the number of generations, and the probabilities of the different operators, as shown in Table~\ref{tab_geneticparameters}.

In addition to the typical parameters of a traditional GA, in a distributed design, it is necessary to configure the distributed execution of the algorithm. Therefore, Table~\ref{tab_geneticparameters} also includes the configuration related to the number of workers, the size of the sub-populations, and the neighborhood radius (the value used to define the neighboring devices in the proposed \textit{Neighbor-aware} design).

Regarding the infrastructure, the fog computing environment where the experiments will be conducted must be defined. This includes specifying the number of nodes, their network topology, the latency of the network, the number of applications, the available resources in the nodes, the resources consumed by the applications, and the users of the system. The specific parameters and their corresponding values are shown in Table~\ref{tab_experimentscenarioparameters}, which have been based on previous works~\cite{ifogsimgupta17,talavera2022geneticbased}.

It is important to note that the number of fog devices may exceed the number of workers. It is not necessary to deploy a worker in each fog device, allowing for a smaller number of workers compared to the total number of fog devices. This flexibility allows for more efficient resource utilization and can be adjusted based on the specific requirements and constraints of the fog computing environment.

\begin{table}[t!]
	\caption{Infrastructure configuration for the experiments.}
	\label{tab_experimentscenarioparameters}
	\centering
	\begin{tabular}{p{4.5cm}|r}
		\toprule
		\textbf{Feature}
		&   \textbf{Value}\\
		\midrule
\multicolumn{2}{l}{\textbf{Infrastructure}}  \\ \midrule
		 Number of fog devices & 100 \\ \midrule
		 Infrastructure topology & Barabasi-Albert \\ \midrule
		 Worker placement criterion & Betweenness centrality\\ \midrule
		 Fog network latency (ms) & [2..6] \\ \midrule
		 Cloud network latency (ms) & 100 \\ \midrule
		 Fog device resources (resource units) & [1..4] \\ \midrule
         Percentage of gateways fog devices (\%) & 25 \\ \midrule
\multicolumn{2}{l}{\textbf{Application}}  \\ \midrule
		 Number of applications & 10 \\ \midrule
		 Application resources (resource units) & [1..2] \\ \midrule
		 
\multicolumn{2}{l}{\textbf{User}}  \\ \midrule
		 User inter-request time (ms) & [5..10] \\ \midrule
         Application popularity per gateway (\%) & [0..75]  \\		
		\bottomrule
	\end{tabular}
\end{table}

By conducting multiple repetitions of the experiments, we can account for the stochastic nature of genetic algorithms and obtain more reliable and statistically significant results. In our study, we have performed 10 repetitions of the experiments for each of the four scenarios: the three distributed designs and the centralized execution of the NSGA-II. This allows us to collect sufficient data to analyze the average values and assess the effectiveness of the proposed designs in a robust manner.

The obtained results from these repetitions will be analyzed and compared in terms of solution quality, convergence, and network usage to evaluate the performance and effectiveness of each design proposal in the fog computing environment.

\subsection{Analysis of the solution quality}

\begin{figure*}[t!]
\centering
\subfloat[Aggregation of the 40 Pareto fronts obtained in the 10 experiment repetitions for each of the 4 experiment scenarios.\label{fig_aggregatedparetos4proposals}]{
\includegraphics[width=0.65\textwidth,trim=20 5 45 40,clip]{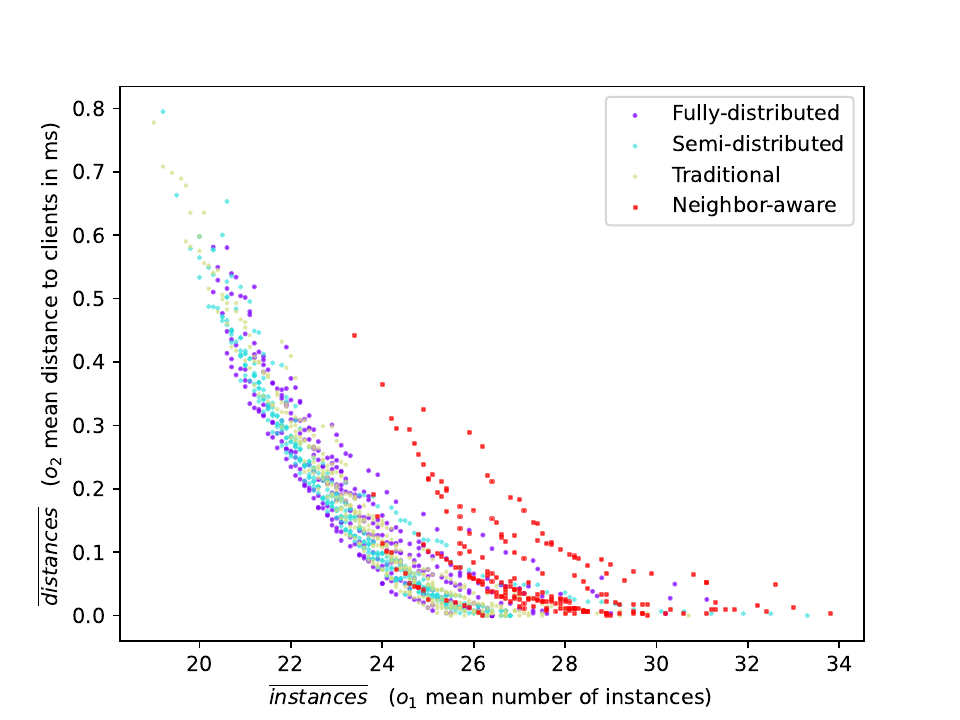}}
\\
\subfloat[Reference Pareto fronts calculated for each experiment scenario.\label{fig_uniqueparetos4proposals}]{
\includegraphics[width=0.65\textwidth,trim=20 5 45 40,clip]{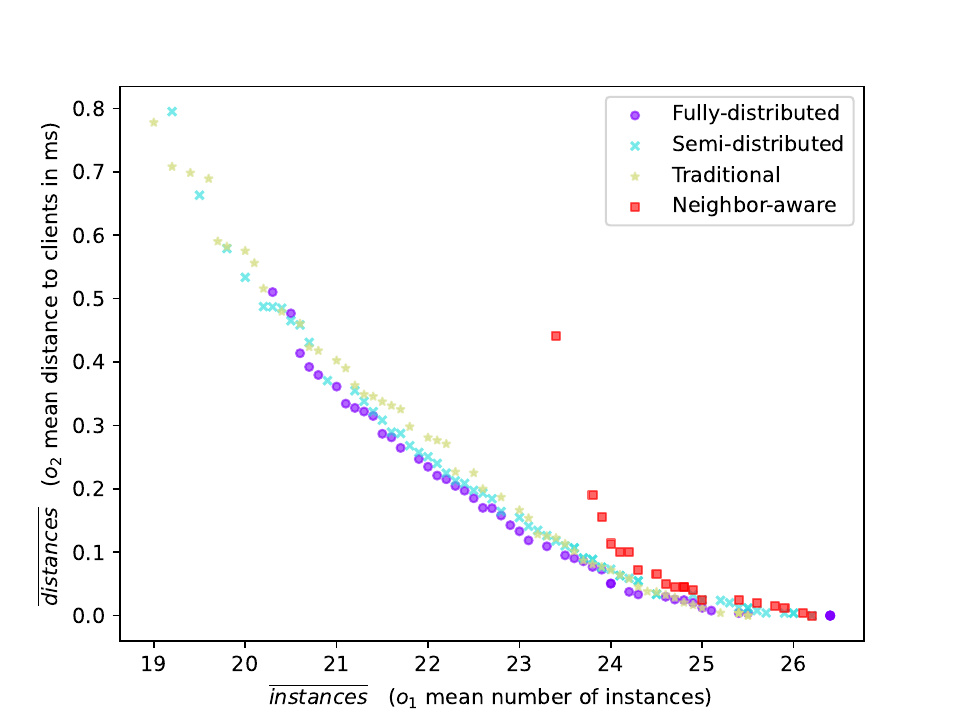}
}
\caption{Objective spaces of the four experiment scenarios.}
\label{fig_paretos4proposals}
\end{figure*}

The first part of the result analysis is devoted to the comparison of the solution quality. By comparing the solution quality across the four experiment scenarios, we can assess the impact of the distributed designs on the optimization process. These distributed designs influence in parent selection, by how the higher probabilities are assigned to the best solutions.

We evaluate the quality of the solutions by comparing the Pareto fronts obtained from each of the proposed distributed designs. Figure~\ref{fig_aggregatedparetos4proposals} presents the 40 Pareto fronts obtained from the 10 experiment repetitions for the 4 scenarios. To facilitate the comparison, we also present the reference Pareto front for each scenario in Figure~\ref{fig_uniqueparetos4proposals}.

The reference front is used in many multi-objective optimization metrics to evaluate how close the obtained Pareto fronts are to a known or approximate optimal solution~\cite{panagant2021comparative}. When the real Pareto cannot be calculated, as is the case we study, the approximate optimal needs to be calculated. Thus, we calculate the reference front as the set of solutions that are not dominated by any other solutions obtained from the compared Pareto fronts. Formally, the reference front of a set of Pareto fronts ($pf_i \in PF$) is obtained by joining all the Pareto fronts ($\bigcup_{i} pf_i$), and calculating the new Pareto front of this set of joined solutions. Consequently, we have joined the Pareto sets obtained in the repetitions of each experiment scenario and the four reference fronts have been obtained and represented in Figure~\ref{fig_uniqueparetos4proposals}.

Upon analyzing Figure~\ref{fig_uniqueparetos4proposals}, it is evident that the \textit{Fully-distributed} design achieves a higher number of solutions that dominate the solutions from the other scenarios. However, the differences in objective values between \textit{Fully-distributed} and the \textit{Semi-distributed} and \textit{Traditional} designs are minimal. In any case, this analysis is slightly different if we consider all the Pareto fronts represented in Figure~\ref{fig_aggregatedparetos4proposals} instead of the reference ones. By observing all the fronts, the \textit{Semi-distributed} design shows an overall highest minimization of the optimization objectives.

\begin{figure} [t!]
\centering
{
\includegraphics[width=0.45\textwidth,trim=20 15 45 40,clip]{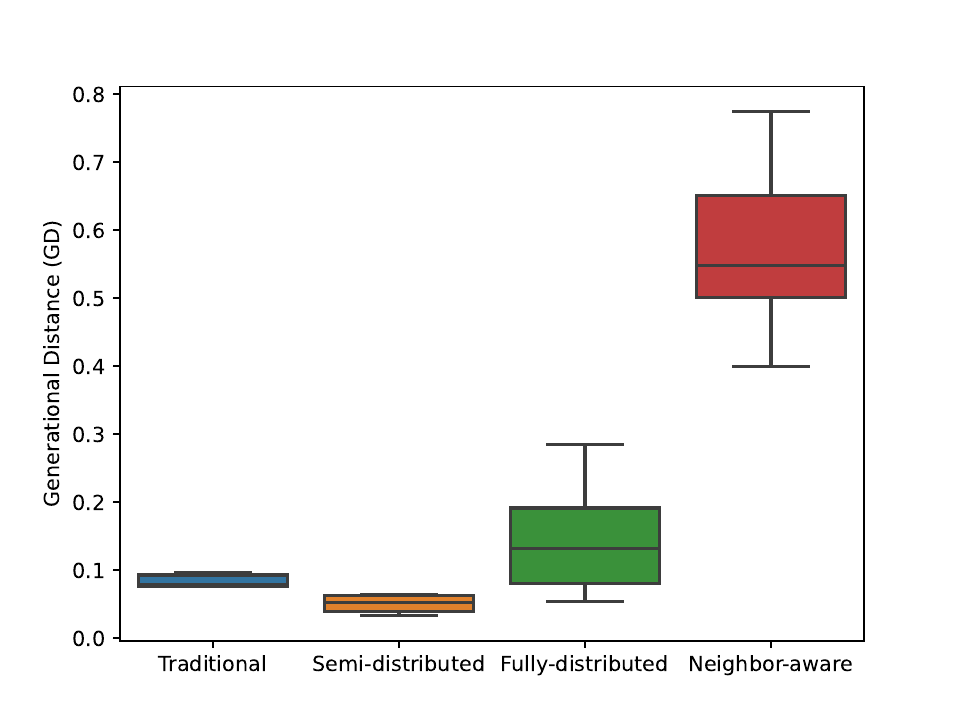}
}
\caption{Quantitative evaluation of the minimization of the objectives of the Pareto fronts through the Generational Distance metric.}
\label{fig_gd}
\end{figure}

A more precise analysis can be performed if we consider some numerical metric to compare the Pareto fronts. Generational Distance (GD) is used to measure distances between an obtained Pareto front and the real (or reference) front~\cite{panagant2021comparative}. A lower value of GD represents a better performance, i.e., a smaller distance to the reference Pareto. Figure~\ref{fig_gd} represents the box plots of the GD for each Pareto front in Figure~\ref{fig_aggregatedparetos4proposals} organized by experiment scenario, and compared to the reference front obtained from the aggregation of the 40 fronts. 

The box plots clearly show that the \textit{Semi-distributed} obtain the best results, the smallest distances, closely followed by {Traditional} with very similar results.

\begin{figure} [t!]
\centering
{
\includegraphics[width=0.45\textwidth,trim=12 15 45 40,clip]{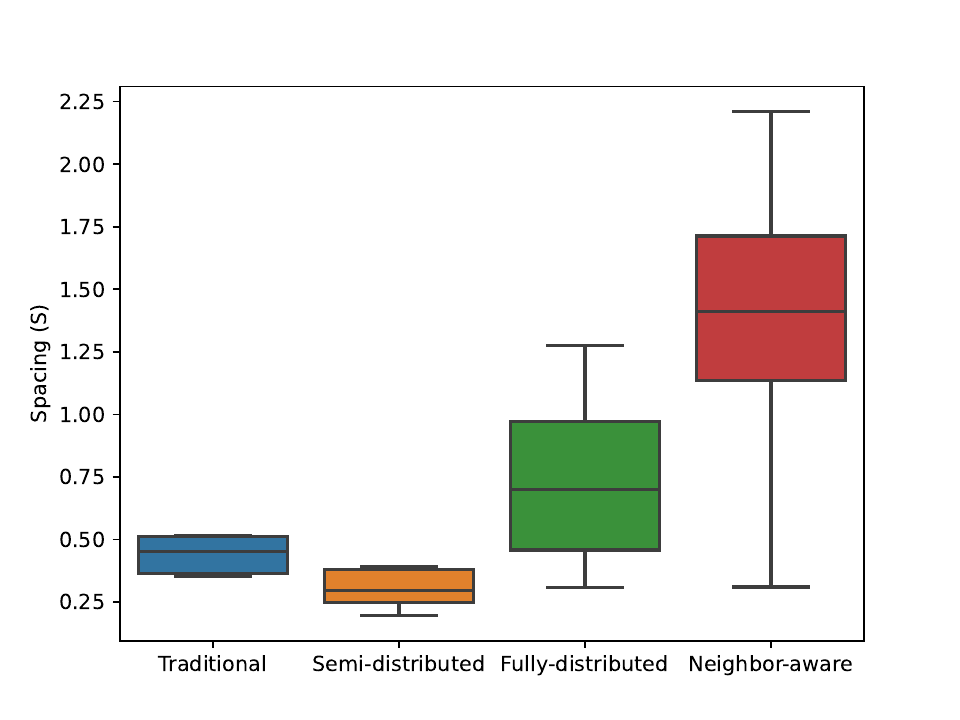}
}
\caption{Quantitative evaluation of the diversity of solutions of the Pareto fronts through the Spacing metric.}
\label{fig_s}
\end{figure}

If we focus the analysis of the quality of the solution in terms of the diversity of the solutions in the Pareto fronts, \textit{Fully-distributed} exhibits a lower diversity compared to {Traditional} and \textit{Semi-distributed}. For example, solutions with a mean number of instances smaller than 20 are exclusively obtained with the two latter designs. Diversity is related to how wide the range of optimized solutions is. A higher diversity offers a greater number of alternatives, with a wider range of optimization values, to choose one specific solution. This analysis can be performed by observing the Pareto fronts (Figure~\ref{fig_aggregatedparetos4proposals}) or numerically with a quality metric. Spacing (S) is a metric to measure how evenly the solutions are distributed along the Pareto front~\cite{6413195}. A lower value reflects a higher diversity in the Pareto.

Figure~\ref{fig_s} represents the box plots of the Spacing of the Pareto fronts organized by experiment scenario. The values of the Spacing confirm the conclusion observed in Figure~\ref{fig_aggregatedparetos4proposals}, because \textit{Traditional} and \textit{Semi-distributed} reflect much smaller values than \textit{Fully-distributed}. And \textit{Neighbor-aware} is clearly the worst case, also in terms of diversity.

Overall, these observations highlight the trade-offs between solution quality, and diversity in the different distributed designs. While \textit{Fully-distributed} achieves a slightly higher number of dominating solutions, it lacks diversity of the solutions.
On the contrary, \textit{Neighbor-aware} is clearly the worst case due to its constrained neighbor selection approach. It is important to carefully consider these factors when selecting a suitable distributed design for fog resource optimization based on the specific requirements and constraints of the application domain. Although these differences could be compensated by the behavior of these designs in terms of network load.

\begin{figure*}[h!]
	\centering

\includegraphics[width=0.52\textwidth,trim=35 5 78 30,clip]{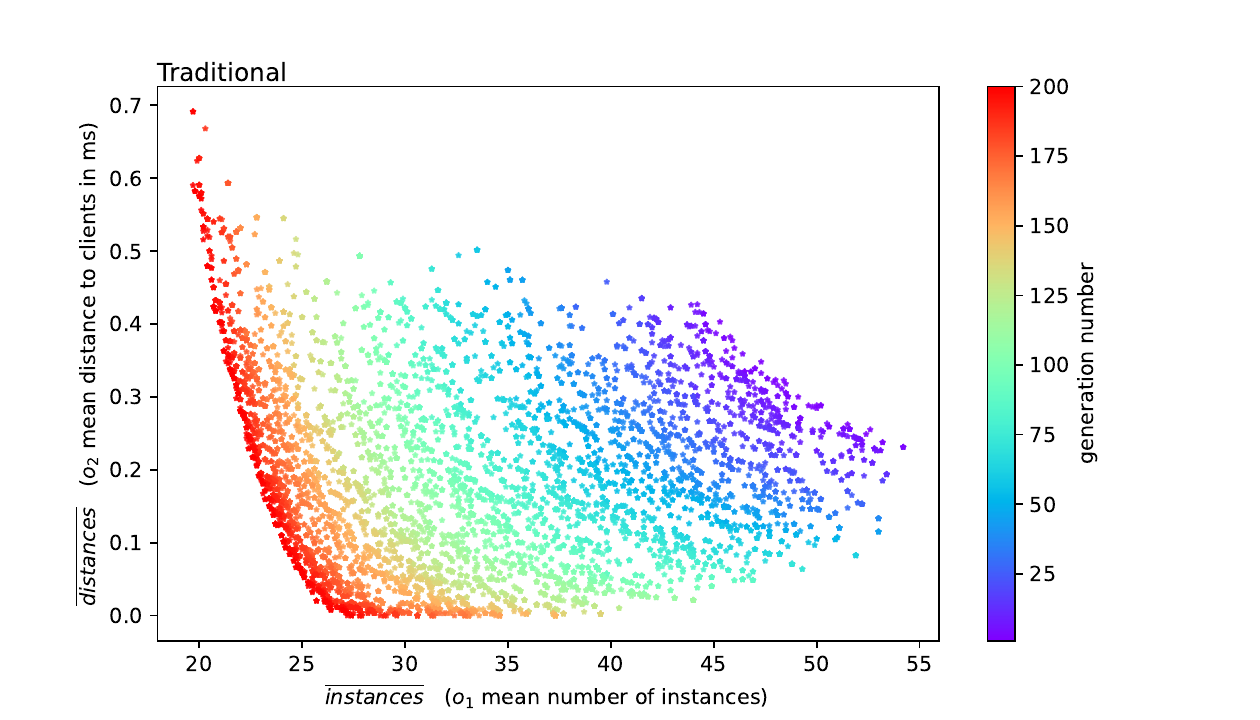}
\hfill
\includegraphics[width=0.455\textwidth,trim=35 5 140 30,clip]{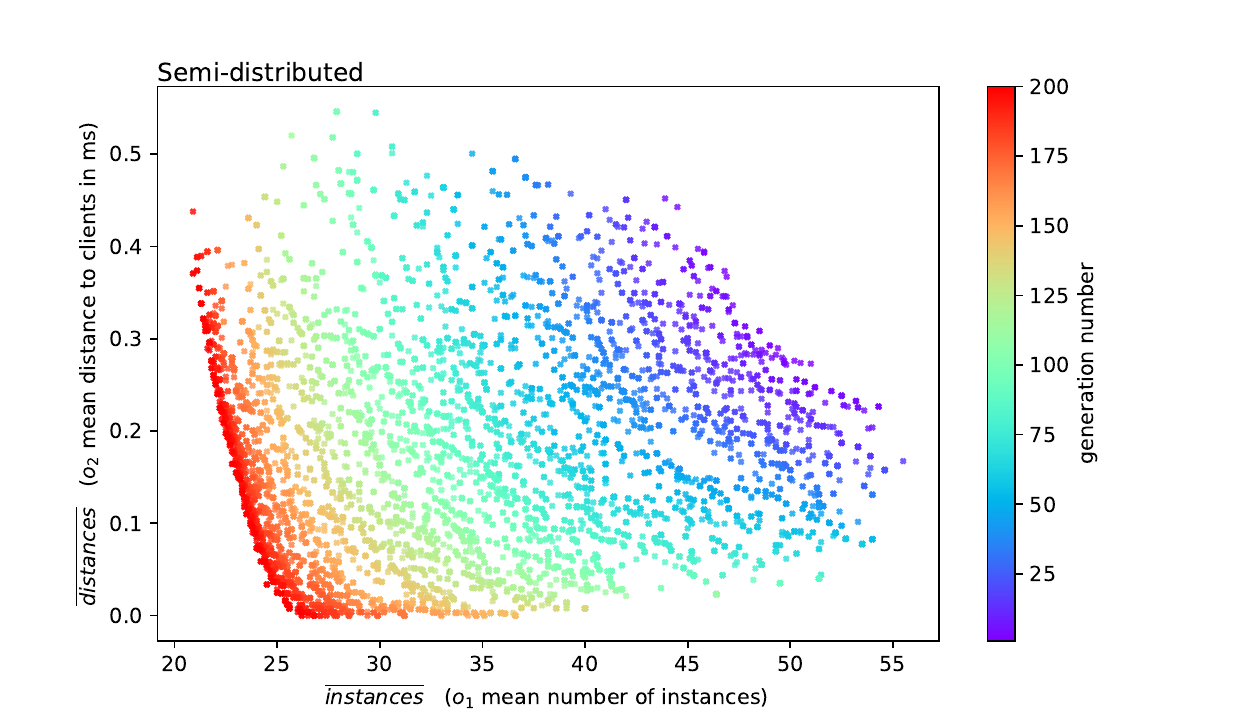}

\includegraphics[width=0.52\textwidth,trim=35 5 78 30,clip]{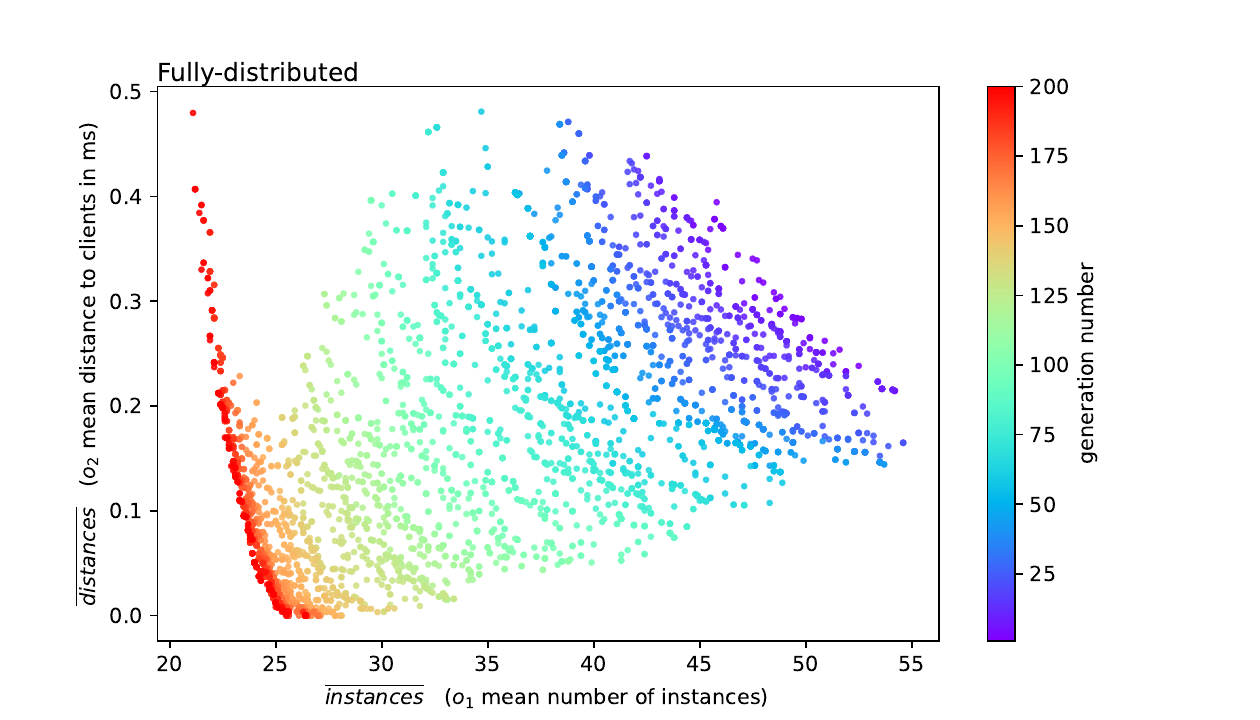}
\hfill
\includegraphics[width=0.455\textwidth,trim=35 5 140 30,clip]{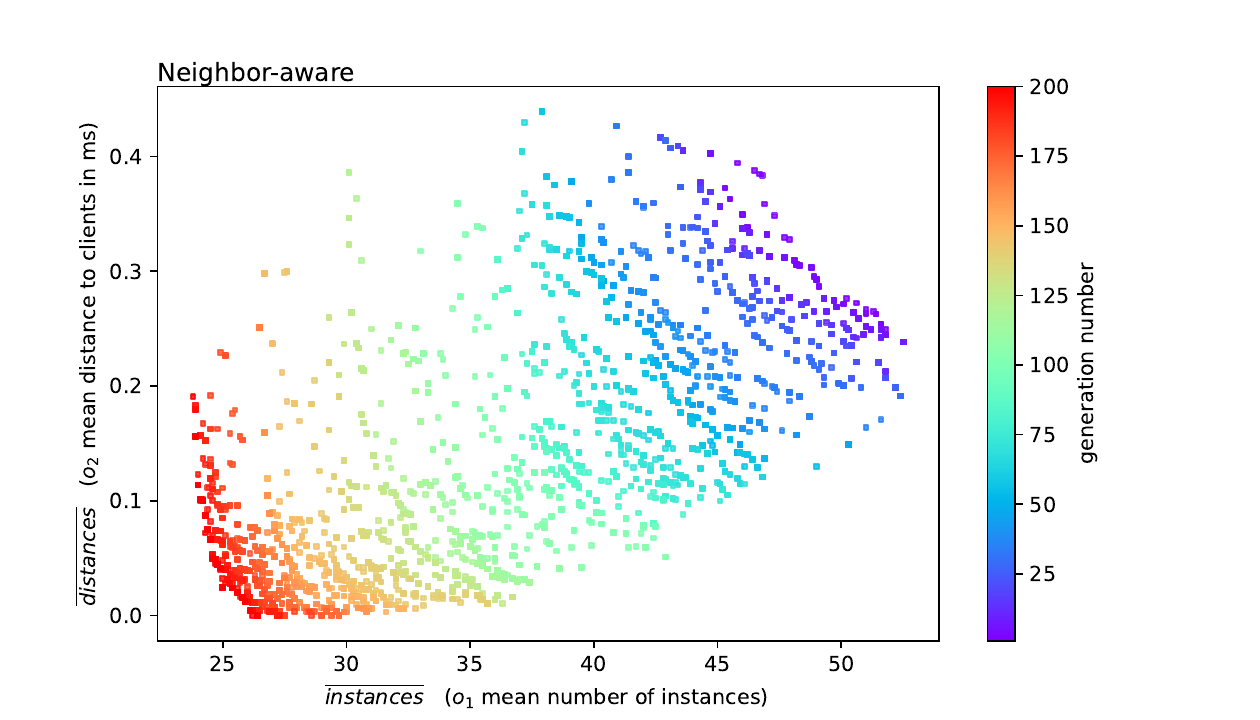}

	\caption{Example of the Pareto evolution for one experiment execution of each scenario.}\label{fig_paretoevolution}
\end{figure*}

The analysis of the evolution of the optimization process is also valuable because we can gain a better understanding of how the optimization process unfolds over time. This evaluation was conducted by examining the changes in the Pareto fronts over generations. With 10 repetitions for each of the four experiment scenarios, a total of 40 plots were generated to illustrate the evolution of the Pareto fronts. From these, we have selected one representative plot for each scenario (Figure~\ref{fig_paretoevolution})\footnote{The remaining 40 plots for each execution are available in the public source code and data repository.}. The trends observed in these selected plots are indicative of the overall behavior observed in all executions, and the conclusions drawn can be generalized to all cases.

In a GA, the differences between the Pareto fronts, in terms of objective value minimization and solution diversity, are typically more pronounced in the initial generations, and the rate of improvement tends to slow down as the generations progress~\cite{safe2004stopping}. The ultimate goal of a GA is for the Pareto front to converge towards the true Pareto front, and as a result, the disparities between the Pareto fronts in the later generations should be minimal. When this convergence is observed, the stopping criterion of the GA can be met, indicating that further generations are unlikely to yield significant improvements.

By analyzing the evolution of the Pareto fronts in the selected experiments in Figure~\ref{fig_paretoevolution}, we observe a consistent pattern across all four scenarios. The Pareto fronts in the final generations, colored in red, exhibit only marginal differences. This indicates that the optimization process has reached a steady state, and the additional generations are unlikely to result in significant improvements. Therefore, we can conclude that the optimization process has effectively converged, and the stopping criteria have been adequately met.









\subsection{Analysis of the convergence between execution repetitions}

The convergence of solutions between repetitions of executions of a GA is of crucial importance in assessing the reliability and consistency of the algorithm's performance. In a stochastic optimization process like a GA, the random operations, such as initial population, parent selection, crossover, etc., can lead to different solutions in each execution repetition. However, if the algorithm is well-designed and effectively explores the search space, it should converge to similar or identical solutions across different repetitions. In our proposals, panmictic nature of the GA is increasingly limited, and the convergence of the solutions between repetitions is required. A high level of convergence indicates that the algorithm is stable and consistently finds good solutions, reducing the influence of this limitation.

We first analyze variances of the values of the GDs. The most destacable results are obtained in \textit{Semi-distributed} and \textit{Traditional}, wich show very small variances in their GDs. The reason for these small variances is that the Pareto fronts obtained in different experiment executions are very similar, i.e., the convergence of the results of these repetitions is very high. This is the desired behavior because a single repetition of the optimization would guarantee suitable results.

Apart from the small variances in the GDs of Figure~\ref{fig_gd}, the convergence of the solutions of our experiments can be assessed by observing the Pareto fronts from each repetition in Figure~\ref{fig_aggregatedparetos4proposals}. 

It is evident that the \textit{Semi-distributed} and \textit{Traditional} designs demonstrate high convergence, with the solutions from different repetitions being closely clustered and concentrated in the same regions of the objective space. This suggests a high level of consistency and reproducibility in the experiments, indicating that these designs are less affected by the stochastic nature of the GA in general, and by the limited selection of the parents of a mating operation. Although not as consistent as the \textit{Semi-distributed} and \textit{Traditional} designs, the \textit{Fully-distributed} design shows potential for achieving reliable and stable results. 

On the contrary, the \textit{Neighbor-aware} design yields the least favorable results, both in terms of objective values and convergence of the results of different repetitions. This is observed clearly in the scatter plots (Figure~\ref{fig_aggregatedparetos4proposals}) as the solutions from different Pareto sets are widely scattered, and they are very far from the non-dominated solutions. Moreover, it is clearly confirmed numerically in Figure~\ref{fig_gd}), where the values of the GDs are much bigger than the other three scenarios and with a much higher variance in these values. This suggests that the constraint of selecting neighbors for solution interchange imposes limitations on the optimization process, leading to suboptimal solutions and a lack of consistency and stability in the optimization process, resulting in significantly varied solutions across repetitions.

\begin{figure} [t!]
\centering
{
\includegraphics[width=0.45\textwidth,trim=0 5 45 40,clip]{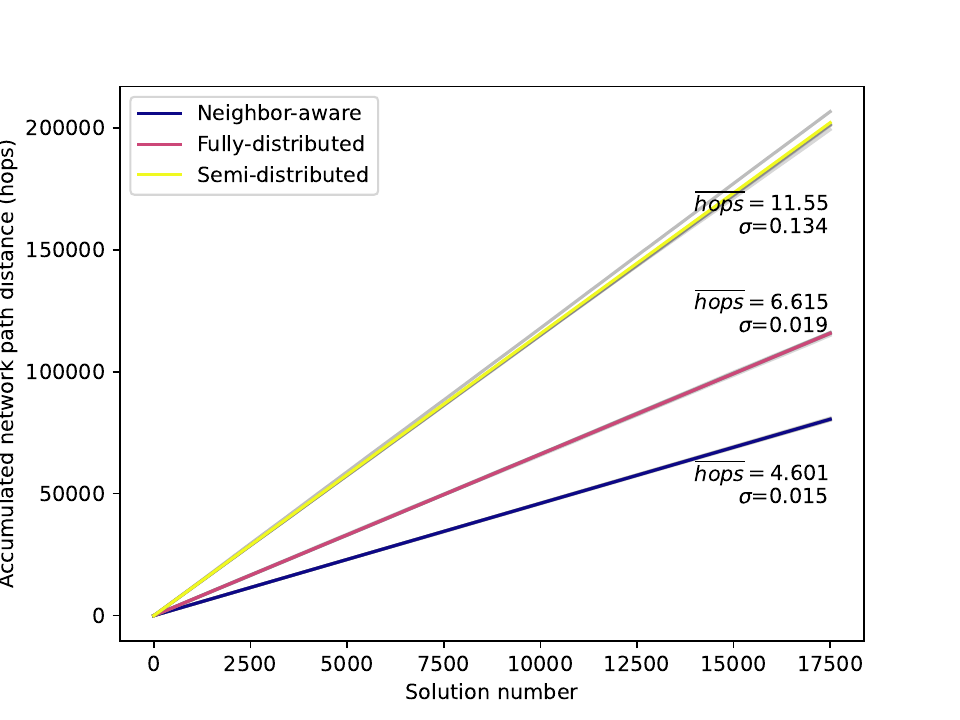}
}
\caption{Accumulated number of hops of the network messages along the generations of the solution.}
\label{fig_accumulativenumberofmessages}
\end{figure}

\subsection{Analysis of the network load}
 
The last part of the result analysis focuses on measuring the network overhead, which is a crucial factor in fog computing due to its highly geographically distributed nature. This analysis has been conducted only for the three distributed experiment scenarios, as the traditional scenario executed on a single computer does not involve inter-node communication. It is worth noting that the number of messages between nodes for each mating operation is always the same, and the variations of the network load are only due to the different number of hops that the messages go through. Remember that in the \textit{Semi-distributed} proposal, the number of messages for each mating operation is three (Figure~\ref{fig_semidistributedmessages}, coordinator to \textit{worker\_m}; \textit{worker\_m} to \textit{worker\_n}; and \textit{worker\_n} to coordinator), and this number is reduced to two in the case of the \textit{Fully-distributed} and \textit{Neighbor-aware} (Figure~\ref{fig_fullydistributedmessages}, \textit{worker\_n} to \textit{worker\_m}; and \textit{worker\_m} to \textit{worker\_n}).

Consequently, we quantified the network overhead as the number of network segments (hops) that the messages go through during the optimization process~\cite{10.1145/2503210.2503263}. Specifically, we have multiplied the number of messages by the number of hops required to deliver each message from the specific source to the specific destination fog device. 

Figure~\ref{fig_accumulativenumberofmessages} presents the cumulative distribution of hops as the number of solutions created in the optimization process increases. The colored lines represent the mean values obtained from all the executions, while the gray lines depict the values obtained in each individual repetition. The deviation between the colored and gray lines is minimal, indicating consistent results across the executions.  This makes the gray lines almost inappreciable, as they are overlapped by the colored lines representing the mean values. For a clearer analysis, the figure also includes the average value and the standard deviation of the number of hops required for a single mating operation.

As expected, the \textit{Neighbor-aware} design exhibits the lowest network workload, with an average of 4,601 hops required for a mating operation. On the other hand, the \textit{Semi-distributed} design generates the highest workload, as each mating operation involves communication between the coordinator, and the worker storing the second parent, the worker computing the new solutions and, again, the coordinator.

The \textit{Fully-distributed} proposal falls between the two extreme cases, but it is much closer to the \textit{Neighbor-aware} design. In fact, the difference in hops between the \textit{Semi-distributed} and \textit{Neighbor-aware} designs is only 2 hops, whereas the difference between the two extreme cases is almost 6 hops. The difference between the \textit{Semi-distributed} and the best-case scenario appears to be manageable, especially considering the benefits it offers in terms of solution quality compared to the \textit{Neighbor-aware} design.

\subsection{Overall analysis}

The \textit{Semi-distributed} design demonstrates comparable solution quality to the centralized version of NSGA-II (\textit{Traditional}). This is achieved by maintaining centralized storage of the objective space but at the expense of increased network usage. On the other hand, the \textit{Fully-distributed} design yields poorer results due to the lack of this global view of the objective space. However, this design significantly reduces the network load. Lastly, the \textit{Neighbor-aware} approach exhibits the lowest solution quality among the distributed proposals but offers the advantage of minimal network load.

In conclusion, both the \textit{Semi-distributed} and \textit{Fully-distributed} designs have proven to be viable distributed proposals for the fog placement optimization problem, leveraging the execution of NSGA-II within the own fog devices. This conclusion is supported by their ability to achieve comparable results to the centralized version of NSGA-II (\textit{Traditional}) in terms of solution quality. However, it is important to note that the \textit{Semi-distributed} design incurs a higher network load.

\section{Conclusions}
\label{sect_conclusions}

In this paper, we have introduced three progressively distributed proposals for implementing genetic-based resource optimizations in fog domains, performing the execution of GAs on the fog infrastructure's own devices. These proposals have been designed with careful consideration of the features of fog infrastructures, including limited resources in devices, heterogeneous and constrained communication networks, and the geographical distribution of fog devices responsible for executing optimization tasks.

Specifically, we have implemented three variations of the NSGA-II algorithm, tailoring them to the specific design characteristics of each distributed proposal. To facilitate efficient communication, we have utilized the MQTT protocol, known for its lightweight characteristics. For the purpose of evaluation, we have selected the service placement problem on fog devices as a benchmark of fog resource optimization. Overall, we conducted experiments using three distinct scenarios: \textit{Semi-distributed}, \textit{Fully-distributed}, and \textit{Neighbor-aware}. Additionally, we included a control scenario employing a centralized version of the NSGA-II (\textit{Traditional}).

In terms of solution quality, we observed that \textit{Fully-distributed} and \textit{Semi-distributed} achieved similar results. The \textit{Fully-distributed} slightly excelled in terms of number of non-dominated solutions, and the \textit{Semi-distributed} in terms of overall optimization of the objectives and the diversity of the solutions. They both obtained competitive results with regard to the control case, the centralized version of the NSGA-II (\textit{Traditional}). However, \textit{Neighbor-aware} exhibited lower solution diversity and obtained the worst results in terms of objective values. 

Regarding network usage, we found that \textit{Neighbor-aware} had the lowest network overhead, while \textit{Semi-distributed} generated the highest workload due to the need to update the centralized objective space. The \textit{Fully-distributed} scenario had intermediate network usage but with significant benefits in terms of solution quality compared to \textit{Neighbor-aware}.

These findings provide system administrators with guidance for selecting the most suitable distributed design, considering their interest in the balance between solution quality and network efficiency in fog computing environments.

These results open future research efforts focused on enhancing the performance of these distributed designs, specifically in terms of improving the solution quality of the \textit{Neighbor-aware} approach. One potential direction is to explore the combination of the \textit{Neighbor-aware} design with mechanisms that facilitate the dissemination of the best solutions among all workers, breaking the frontier of the neighborhood of a worker for this sub-set of the best solutions. Increasing the exchange of solutions could lead to an enhancement in solution quality, but paying attention to ensure that the network load remains low.

Another promising avenue for future research is to explore the incorporation of quality indicators for the sub-populations in the \textit{Fully-distributed} and \textit{Neighbor-aware} designs. These indicators would provide a measure of the quality of the sub-populations stored in each worker. By integrating these indicators into the parent selection process, a higher probability can be assigned to select workers with superior sub-populations as parents, rather than selecting uniformly from all nodes/neighbors. This approach would increase the likelihood of selecting the best solutions as parents for the mating operation, ultimately enhancing the overall quality of the optimized population.

Finally, an intriguing direction for future research is to explore the integration of different optimization algorithms within distributed GA designs. This hybrid approach could involve employing different meta-heuristics in the workers. By incorporating solution interchange between the different algorithms during each optimization step or mating operation, it would be possible to leverage the strengths of each algorithm and potentially achieve superior optimization performance.


\section*{Funding}
Funding: This work was supported by MCIN/AEI/10.13039/501100011033 [grant number PID2021-128071OB-I00] and by the European Union NextGenerationEU/PRTR [grant number PID2021-128071OB-I00].

\section*{Data Availability}

The data generated in this study and the source code of the
scripts are public available in the repository: \\ \url{https://github.com/acsicuib/GAmqtt}





\bibliographystyle{elsarticle-num-names} 

\bibliography{elsarticle-template-num}

\begin{thebibliography}{40}
\providecommand{\natexlab}[1]{#1}
\providecommand{\url}[1]{\texttt{#1}}
\providecommand{\urlprefix}{URL }
\expandafter\ifx\csname urlstyle\endcsname\relax
  \providecommand{\doi}[1]{doi:\discretionary{}{}{}#1}\else
  \providecommand{\doi}[1]{doi:\discretionary{}{}{}\begingroup
  \urlstyle{rm}\url{#1}\endgroup}\fi
\providecommand{\bibinfo}[2]{#2}

\bibitem[{Srirama et~al.(2021)Srirama, Dick, and Adhikari}]{SRIRAMA2021439}
\bibinfo{author}{S.~N. Srirama}, \bibinfo{author}{F.~M.~S. Dick},
  \bibinfo{author}{M.~Adhikari}, \bibinfo{title}{Akka framework based on the
  Actor model for executing distributed Fog Computing applications},
  \bibinfo{journal}{Future Generation Computer Systems} \bibinfo{volume}{117}
  (\bibinfo{year}{2021}) \bibinfo{pages}{439--452}, ISSN
  \bibinfo{issn}{0167-739X},
  \doi{\bibinfo{doi}{https://doi.org/10.1016/j.future.2020.12.011}},
  \urlprefix\url{https://www.sciencedirect.com/science/article/pii/S0167739X20330739}.

\bibitem[{Guerrero et~al.(2022)Guerrero, Lera, and Juiz}]{GUERRERO2022101094}
\bibinfo{author}{C.~Guerrero}, \bibinfo{author}{I.~Lera},
  \bibinfo{author}{C.~Juiz}, \bibinfo{title}{Genetic-based optimization in fog
  computing: Current trends and research opportunities},
  \bibinfo{journal}{Swarm and Evolutionary Computation} \bibinfo{volume}{72}
  (\bibinfo{year}{2022}) \bibinfo{pages}{101094}, ISSN
  \bibinfo{issn}{2210-6502},
  \doi{\bibinfo{doi}{https://doi.org/10.1016/j.swevo.2022.101094}},
  \urlprefix\url{https://www.sciencedirect.com/science/article/pii/S2210650222000645}.

\bibitem[{Morell and Alba(2018)}]{10.1007/978-3-030-00374-6_24}
\bibinfo{author}{J.~{\'A}. Morell}, \bibinfo{author}{E.~Alba},
  \bibinfo{title}{Running Genetic Algorithms in the Edge: A First Analysis},
  in: \bibinfo{editor}{F.~Herrera}, \bibinfo{editor}{S.~Damas},
  \bibinfo{editor}{R.~Montes}, \bibinfo{editor}{S.~Alonso},
  \bibinfo{editor}{{\'O}.~Cord{\'o}n}, \bibinfo{editor}{A.~Gonz{\'a}lez},
  \bibinfo{editor}{A.~Troncoso} (Eds.), \bibinfo{booktitle}{Advances in
  Artificial Intelligence}, \bibinfo{publisher}{Springer International
  Publishing}, \bibinfo{address}{Cham}, ISBN \bibinfo{isbn}{978-3-030-00374-6},
  \bibinfo{pages}{251--261}, \bibinfo{year}{2018}.

\bibitem[{Moysiadis et~al.(2018)Moysiadis, Sarigiannidis, and
  Moscholios}]{moysiadis2018}
\bibinfo{author}{V.~Moysiadis}, \bibinfo{author}{P.~Sarigiannidis},
  \bibinfo{author}{I.~Moscholios}, \bibinfo{title}{{Towards Distributed Data
  Management in Fog Computing}}, \bibinfo{journal}{{Wireless Communications and
  Mobile Computing}} \bibinfo{volume}{2018} (\bibinfo{year}{2018})
  \bibinfo{pages}{14}, \urlprefix\url{10.1155/2018/7597686}.

\bibitem[{Deb et~al.(2002)Deb, Pratap, Agarwal, and Meyarivan}]{deb2002fast}
\bibinfo{author}{K.~Deb}, \bibinfo{author}{A.~Pratap},
  \bibinfo{author}{S.~Agarwal}, \bibinfo{author}{T.~Meyarivan},
  \bibinfo{title}{A fast and elitist multiobjective genetic algorithm:
  NSGA-II}, \bibinfo{journal}{IEEE transactions on evolutionary computation}
  \bibinfo{volume}{6}~(\bibinfo{number}{2}) (\bibinfo{year}{2002})
  \bibinfo{pages}{182--197}.

\bibitem[{Zitzler et~al.(2001)Zitzler, Laumanns, and Thiele}]{zitzler2001spea2}
\bibinfo{author}{E.~Zitzler}, \bibinfo{author}{M.~Laumanns},
  \bibinfo{author}{L.~Thiele}, \bibinfo{title}{SPEA2: Improving the strength
  Pareto evolutionary algorithm}, \bibinfo{journal}{TIK-report}
  \bibinfo{volume}{103}.

\bibitem[{Von~L{\"u}cken et~al.(2014)Von~L{\"u}cken, Bar{\'a}n, and
  Brizuela}]{von2014survey}
\bibinfo{author}{C.~Von~L{\"u}cken}, \bibinfo{author}{B.~Bar{\'a}n},
  \bibinfo{author}{C.~Brizuela}, \bibinfo{title}{A survey on multi-objective
  evolutionary algorithms for many-objective problems},
  \bibinfo{journal}{Computational optimization and applications}
  \bibinfo{volume}{58} (\bibinfo{year}{2014}) \bibinfo{pages}{707--756}.

\bibitem[{Ghobaei-Arani et~al.(2020)Ghobaei-Arani, Souri, and
  Rahmanian}]{ghobaei2020resource}
\bibinfo{author}{M.~Ghobaei-Arani}, \bibinfo{author}{A.~Souri},
  \bibinfo{author}{A.~A. Rahmanian}, \bibinfo{title}{Resource management
  approaches in fog computing: a comprehensive review},
  \bibinfo{journal}{Journal of Grid Computing}
  \bibinfo{volume}{18}~(\bibinfo{number}{1}) (\bibinfo{year}{2020})
  \bibinfo{pages}{1--42}.

\bibitem[{Martinez et~al.(2021)Martinez, Hafid, and Jarray}]{9194714}
\bibinfo{author}{I.~Martinez}, \bibinfo{author}{A.~S. Hafid},
  \bibinfo{author}{A.~Jarray}, \bibinfo{title}{Design, Resource Management, and
  Evaluation of Fog Computing Systems: A Survey}, \bibinfo{journal}{IEEE
  Internet of Things Journal} \bibinfo{volume}{8}~(\bibinfo{number}{4})
  (\bibinfo{year}{2021}) \bibinfo{pages}{2494--2516},
  \doi{\bibinfo{doi}{10.1109/JIOT.2020.3022699}}.

\bibitem[{Hong and Varghese(2019)}]{10.1145/3326066}
\bibinfo{author}{C.-H. Hong}, \bibinfo{author}{B.~Varghese},
  \bibinfo{title}{Resource Management in Fog/Edge Computing: A Survey on
  Architectures, Infrastructure, and Algorithms}, \bibinfo{journal}{ACM Comput.
  Surv.} \bibinfo{volume}{52}~(\bibinfo{number}{5}), ISSN
  \bibinfo{issn}{0360-0300}, \doi{\bibinfo{doi}{10.1145/3326066}},
  \urlprefix\url{https://doi.org/10.1145/3326066}.

\bibitem[{{Mennes} et~al.(2016){Mennes}, {Spinnewyn}, {Latré}, and
  {Botero}}]{7776569}
\bibinfo{author}{R.~{Mennes}}, \bibinfo{author}{B.~{Spinnewyn}},
  \bibinfo{author}{S.~{Latré}}, \bibinfo{author}{J.~F. {Botero}},
  \bibinfo{title}{GRECO: A Distributed Genetic Algorithm for Reliable
  Application Placement in Hybrid Clouds}, in: \bibinfo{booktitle}{2016 5th
  IEEE International Conference on Cloud Networking (Cloudnet)},
  \bibinfo{pages}{14--20}, \doi{\bibinfo{doi}{10.1109/CloudNet.2016.45}},
  \bibinfo{year}{2016}.

\bibitem[{Yang et~al.(2020)Yang, Wen, McKee, Lin, Xu, and
  Garraghan}]{9c434213a47d45b486ebc9c416d132d0}
\bibinfo{author}{R.~Yang}, \bibinfo{author}{Z.~Wen},
  \bibinfo{author}{D.~McKee}, \bibinfo{author}{T.~Lin},
  \bibinfo{author}{J.~Xu}, \bibinfo{author}{P.~Garraghan}, \bibinfo{title}{Fog
  Orchestration and Simulation for IoT Services}, \bibinfo{publisher}{Wiley},
  ISBN \bibinfo{isbn}{1119501091}, \bibinfo{pages}{179--212},
  \bibinfo{year}{2020}.

\bibitem[{{Wen} et~al.(2017){Wen}, {Yang}, {Garraghan}, {Lin}, {Xu}, and
  {Rovatsos}}]{7867735}
\bibinfo{author}{Z.~{Wen}}, \bibinfo{author}{R.~{Yang}},
  \bibinfo{author}{P.~{Garraghan}}, \bibinfo{author}{T.~{Lin}},
  \bibinfo{author}{J.~{Xu}}, \bibinfo{author}{M.~{Rovatsos}},
  \bibinfo{title}{Fog Orchestration for Internet of Things Services},
  \bibinfo{journal}{IEEE Internet Computing}
  \bibinfo{volume}{21}~(\bibinfo{number}{2}) (\bibinfo{year}{2017})
  \bibinfo{pages}{16--24}, \doi{\bibinfo{doi}{10.1109/MIC.2017.36}}.

\bibitem[{Cardellini et~al.(2015)Cardellini, Grassi, Presti, and
  Nardelli}]{7405527}
\bibinfo{author}{V.~Cardellini}, \bibinfo{author}{V.~Grassi},
  \bibinfo{author}{F.~L. Presti}, \bibinfo{author}{M.~Nardelli},
  \bibinfo{title}{On QoS-aware scheduling of data stream applications over fog
  computing infrastructures}, in: \bibinfo{booktitle}{2015 IEEE Symposium on
  Computers and Communication (ISCC)}, \bibinfo{pages}{271--276},
  \doi{\bibinfo{doi}{10.1109/ISCC.2015.7405527}}, \bibinfo{year}{2015}.

\bibitem[{Donassolo et~al.(2019)Donassolo, Fajjari, Legrand, and
  Mertikopoulos}]{8651835}
\bibinfo{author}{B.~Donassolo}, \bibinfo{author}{I.~Fajjari},
  \bibinfo{author}{A.~Legrand}, \bibinfo{author}{P.~Mertikopoulos},
  \bibinfo{title}{Fog Based Framework for IoT Service Provisioning}, in:
  \bibinfo{booktitle}{2019 16th IEEE Annual Consumer Communications \&
  Networking Conference (CCNC)}, \bibinfo{pages}{1--6},
  \doi{\bibinfo{doi}{10.1109/CCNC.2019.8651835}}, \bibinfo{year}{2019}.

\bibitem[{Forti et~al.(2021)Forti, Gaglianese, and Brogi}]{FORTI2021605}
\bibinfo{author}{S.~Forti}, \bibinfo{author}{M.~Gaglianese},
  \bibinfo{author}{A.~Brogi}, \bibinfo{title}{Lightweight self-organising
  distributed monitoring of Fog infrastructures}, \bibinfo{journal}{Future
  Generation Computer Systems} \bibinfo{volume}{114} (\bibinfo{year}{2021})
  \bibinfo{pages}{605--618}, ISSN \bibinfo{issn}{0167-739X},
  \doi{\bibinfo{doi}{https://doi.org/10.1016/j.future.2020.08.011}},
  \urlprefix\url{https://www.sciencedirect.com/science/article/pii/S0167739X19334582}.

\bibitem[{Harada and Alba(2020)}]{10.1145/3400031}
\bibinfo{author}{T.~Harada}, \bibinfo{author}{E.~Alba},
  \bibinfo{title}{Parallel Genetic Algorithms: A Useful Survey},
  \bibinfo{journal}{ACM Comput. Surv.}
  \bibinfo{volume}{53}~(\bibinfo{number}{4}), ISSN \bibinfo{issn}{0360-0300},
  \doi{\bibinfo{doi}{10.1145/3400031}},
  \urlprefix\url{https://doi.org/10.1145/3400031}.

\bibitem[{Puliafito et~al.(2019)Puliafito, Mingozzi, Longo, Puliafito, and
  Rana}]{10.1145/3301443}
\bibinfo{author}{C.~Puliafito}, \bibinfo{author}{E.~Mingozzi},
  \bibinfo{author}{F.~Longo}, \bibinfo{author}{A.~Puliafito},
  \bibinfo{author}{O.~Rana}, \bibinfo{title}{Fog Computing for the Internet of
  Things: A Survey}, \bibinfo{journal}{ACM Trans. Internet Technol.}
  \bibinfo{volume}{19}~(\bibinfo{number}{2}), ISSN \bibinfo{issn}{1533-5399},
  \doi{\bibinfo{doi}{10.1145/3301443}},
  \urlprefix\url{https://doi.org/10.1145/3301443}.

\bibitem[{Naha et~al.(2018)Naha, Garg, Georgakopoulos, Jayaraman, Gao, Xiang,
  and Ranjan}]{8444370}
\bibinfo{author}{R.~K. Naha}, \bibinfo{author}{S.~Garg},
  \bibinfo{author}{D.~Georgakopoulos}, \bibinfo{author}{P.~P. Jayaraman},
  \bibinfo{author}{L.~Gao}, \bibinfo{author}{Y.~Xiang},
  \bibinfo{author}{R.~Ranjan}, \bibinfo{title}{Fog Computing: Survey of Trends,
  Architectures, Requirements, and Research Directions}, \bibinfo{journal}{IEEE
  Access} \bibinfo{volume}{6} (\bibinfo{year}{2018})
  \bibinfo{pages}{47980--48009},
  \doi{\bibinfo{doi}{10.1109/ACCESS.2018.2866491}}.

\bibitem[{Bittencourt et~al.(2018)Bittencourt, Immich, Sakellariou, Fonseca,
  Madeira, Curado, Villas, DaSilva, Lee, and Rana}]{BITTENCOURT2018134}
\bibinfo{author}{L.~Bittencourt}, \bibinfo{author}{R.~Immich},
  \bibinfo{author}{R.~Sakellariou}, \bibinfo{author}{N.~Fonseca},
  \bibinfo{author}{E.~Madeira}, \bibinfo{author}{M.~Curado},
  \bibinfo{author}{L.~Villas}, \bibinfo{author}{L.~DaSilva},
  \bibinfo{author}{C.~Lee}, \bibinfo{author}{O.~Rana}, \bibinfo{title}{The
  Internet of Things, Fog and Cloud continuum: Integration and challenges},
  \bibinfo{journal}{Internet of Things} \bibinfo{volume}{3-4}
  (\bibinfo{year}{2018}) \bibinfo{pages}{134--155}, ISSN
  \bibinfo{issn}{2542-6605},
  \doi{\bibinfo{doi}{https://doi.org/10.1016/j.iot.2018.09.005}},
  \urlprefix\url{https://www.sciencedirect.com/science/article/pii/S2542660518300635}.

\bibitem[{Cant{\'u}-Paz(1998)}]{cantu1998survey}
\bibinfo{author}{E.~Cant{\'u}-Paz}, \bibinfo{title}{A survey of parallel
  genetic algorithms}, \bibinfo{journal}{Calculateurs paralleles, reseaux et
  systems repartis} \bibinfo{volume}{10}~(\bibinfo{number}{2})
  (\bibinfo{year}{1998}) \bibinfo{pages}{141--171}.

\bibitem[{Alba(2005)}]{alba2005parallel}
\bibinfo{author}{E.~Alba}, \bibinfo{title}{Parallel metaheuristics: a new class
  of algorithms}, \bibinfo{publisher}{John Wiley \& Sons},
  \bibinfo{year}{2005}.

\bibitem[{Talbi and Hasle(2013)}]{talbi2013metaheuristics}
\bibinfo{author}{E.-G. Talbi}, \bibinfo{author}{G.~Hasle},
  \bibinfo{title}{Metaheuristics on gpus}, \bibinfo{journal}{J. Parallel
  Distributed Comput.} \bibinfo{volume}{73}~(\bibinfo{number}{1})
  (\bibinfo{year}{2013}) \bibinfo{pages}{1--3}.

\bibitem[{Alba et~al.(1999)Alba, Troya et~al.}]{alba1999survey}
\bibinfo{author}{E.~Alba}, \bibinfo{author}{J.~M. Troya}, et~al.,
  \bibinfo{title}{A survey of parallel distributed genetic algorithms},
  \bibinfo{journal}{Complexity} \bibinfo{volume}{4}~(\bibinfo{number}{4})
  (\bibinfo{year}{1999}) \bibinfo{pages}{31--52}.

\bibitem[{Alba and Tomassini(2002)}]{1041554}
\bibinfo{author}{E.~Alba}, \bibinfo{author}{M.~Tomassini},
  \bibinfo{title}{Parallelism and evolutionary algorithms},
  \bibinfo{journal}{IEEE Transactions on Evolutionary Computation}
  \bibinfo{volume}{6}~(\bibinfo{number}{5}) (\bibinfo{year}{2002})
  \bibinfo{pages}{443--462}, \doi{\bibinfo{doi}{10.1109/TEVC.2002.800880}}.

\bibitem[{Sudholt(2015)}]{Sudholt2015}
\bibinfo{author}{D.~Sudholt}, \bibinfo{title}{Parallel Evolutionary
  Algorithms}, \bibinfo{publisher}{Springer Berlin Heidelberg},
  \bibinfo{address}{Berlin, Heidelberg}, ISBN
  \bibinfo{isbn}{978-3-662-43505-2}, \bibinfo{pages}{929--959},
  \doi{\bibinfo{doi}{10.1007/978-3-662-43505-2_46}},
  \urlprefix\url{https://doi.org/10.1007/978-3-662-43505-2_46},
  \bibinfo{year}{2015}.

\bibitem[{Alba and Troya(2000)}]{alba2000influence}
\bibinfo{author}{E.~Alba}, \bibinfo{author}{J.~M. Troya},
  \bibinfo{title}{Influence of the migration policy in parallel distributed GAs
  with structured and panmictic populations}, \bibinfo{journal}{Applied
  Intelligence} \bibinfo{volume}{12} (\bibinfo{year}{2000})
  \bibinfo{pages}{163--181}.

\bibitem[{Bellavista et~al.(2019)Bellavista, Foschini, Ghiselli, and
  Reale}]{8969615}
\bibinfo{author}{P.~Bellavista}, \bibinfo{author}{L.~Foschini},
  \bibinfo{author}{N.~Ghiselli}, \bibinfo{author}{A.~Reale},
  \bibinfo{title}{MQTT-based Middleware for Container Support in Fog Computing
  Environments}, in: \bibinfo{booktitle}{2019 IEEE Symposium on Computers and
  Communications (ISCC)}, \bibinfo{pages}{1--7},
  \doi{\bibinfo{doi}{10.1109/ISCC47284.2019.8969615}}, \bibinfo{year}{2019}.

\bibitem[{Syswerda(1991)}]{SYSWERDA199194}
\bibinfo{author}{G.~Syswerda}, \bibinfo{title}{A Study of Reproduction in
  Generational and Steady-State Genetic Algorithms}, vol.~\bibinfo{volume}{1}
  of \emph{\bibinfo{series}{Foundations of Genetic Algorithms}},
  \bibinfo{publisher}{Elsevier}, \bibinfo{pages}{94--101},
  \doi{\bibinfo{doi}{https://doi.org/10.1016/B978-0-08-050684-5.50009-4}},
  \urlprefix\url{https://www.sciencedirect.com/science/article/pii/B9780080506845500094},
  \bibinfo{year}{1991}.

\bibitem[{Das and Inuwa(2023)}]{DAS2023100049}
\bibinfo{author}{R.~Das}, \bibinfo{author}{M.~M. Inuwa}, \bibinfo{title}{A
  review on fog computing: Issues, characteristics, challenges, and potential
  applications}, \bibinfo{journal}{Telematics and Informatics Reports}
  \bibinfo{volume}{10} (\bibinfo{year}{2023}) \bibinfo{pages}{100049}, ISSN
  \bibinfo{issn}{2772-5030},
  \doi{\bibinfo{doi}{https://doi.org/10.1016/j.teler.2023.100049}},
  \urlprefix\url{https://www.sciencedirect.com/science/article/pii/S2772503023000099}.

\bibitem[{Brogi et~al.(2020)Brogi, Forti, Guerrero, and Lera}]{brogi2020place}
\bibinfo{author}{A.~Brogi}, \bibinfo{author}{S.~Forti},
  \bibinfo{author}{C.~Guerrero}, \bibinfo{author}{I.~Lera}, \bibinfo{title}{How
  to place your apps in the fog: State of the art and open challenges},
  \bibinfo{journal}{Software: Practice and Experience}
  \bibinfo{volume}{50}~(\bibinfo{number}{5}) (\bibinfo{year}{2020})
  \bibinfo{pages}{719--740}.

\bibitem[{Guerrero et~al.(2019)Guerrero, Lera, and Juiz}]{GUERRERO2019131}
\bibinfo{author}{C.~Guerrero}, \bibinfo{author}{I.~Lera},
  \bibinfo{author}{C.~Juiz}, \bibinfo{title}{Evaluation and efficiency
  comparison of evolutionary algorithms for service placement optimization in
  fog architectures}, \bibinfo{journal}{Future Generation Computer Systems}
  \bibinfo{volume}{97} (\bibinfo{year}{2019}) \bibinfo{pages}{131--144}, ISSN
  \bibinfo{issn}{0167-739X},
  \doi{\bibinfo{doi}{https://doi.org/10.1016/j.future.2019.02.056}},
  \urlprefix\url{https://www.sciencedirect.com/science/article/pii/S0167739X18325147}.

\bibitem[{Birattari and Kacprzyk(2009)}]{birattari2009tuning}
\bibinfo{author}{M.~Birattari}, \bibinfo{author}{J.~Kacprzyk},
  \bibinfo{title}{Tuning metaheuristics: a machine learning perspective}, vol.
  \bibinfo{volume}{197}, \bibinfo{publisher}{Springer}, \bibinfo{year}{2009}.

\bibitem[{Mosayebi and Sodhi(2020)}]{10.1145/3377929.3398136}
\bibinfo{author}{M.~Mosayebi}, \bibinfo{author}{M.~Sodhi},
  \bibinfo{title}{Tuning Genetic Algorithm Parameters Using Design of
  Experiments}, \bibinfo{publisher}{Association for Computing Machinery},
  \bibinfo{address}{New York, NY, USA}, ISBN \bibinfo{isbn}{9781450371278},
  \bibinfo{pages}{1937–1944},
  \urlprefix\url{https://doi.org/10.1145/3377929.3398136},
  \bibinfo{year}{2020}.

\bibitem[{Gupta et~al.(2017)Gupta, Vahid Dastjerdi, Ghosh, and
  Buyya}]{ifogsimgupta17}
\bibinfo{author}{H.~Gupta}, \bibinfo{author}{A.~Vahid Dastjerdi},
  \bibinfo{author}{S.~K. Ghosh}, \bibinfo{author}{R.~Buyya},
  \bibinfo{title}{iFogSim: A toolkit for modeling and simulation of resource
  management techniques in the Internet of Things, Edge and Fog computing
  environments}, \bibinfo{journal}{Software: Practice and Experience}
  \bibinfo{volume}{47}~(\bibinfo{number}{9}) (\bibinfo{year}{2017})
  \bibinfo{pages}{1275--1296},
  \doi{\bibinfo{doi}{https://doi.org/10.1002/spe.2509}}.

\bibitem[{Talavera et~al.(2022)Talavera, Lera, Juiz, and
  Guerrero}]{talavera2022geneticbased}
\bibinfo{author}{F.~Talavera}, \bibinfo{author}{I.~Lera},
  \bibinfo{author}{C.~Juiz}, \bibinfo{author}{C.~Guerrero},
  \bibinfo{title}{Genetic-based fog colony optimization hybridized with
  hierarchical clustering and its influence in the placement of fog services},
  \bibinfo{year}{2022}.

\bibitem[{Panagant et~al.(2021)Panagant, Pholdee, Bureerat, Yildiz, and
  Mirjalili}]{panagant2021comparative}
\bibinfo{author}{N.~Panagant}, \bibinfo{author}{N.~Pholdee},
  \bibinfo{author}{S.~Bureerat}, \bibinfo{author}{A.~R. Yildiz},
  \bibinfo{author}{S.~Mirjalili}, \bibinfo{title}{A comparative study of recent
  multi-objective metaheuristics for solving constrained truss optimisation
  problems}, \bibinfo{journal}{Archives of Computational Methods in
  Engineering}  (\bibinfo{year}{2021}) \bibinfo{pages}{1--17}.

\bibitem[{Yen and He(2014)}]{6413195}
\bibinfo{author}{G.~G. Yen}, \bibinfo{author}{Z.~He},
  \bibinfo{title}{Performance Metric Ensemble for Multiobjective Evolutionary
  Algorithms}, \bibinfo{journal}{IEEE Transactions on Evolutionary Computation}
  \bibinfo{volume}{18}~(\bibinfo{number}{1}) (\bibinfo{year}{2014})
  \bibinfo{pages}{131--144}, \doi{\bibinfo{doi}{10.1109/TEVC.2013.2240687}}.

\bibitem[{Safe et~al.(2004)Safe, Carballido, Ponzoni, and
  Brignole}]{safe2004stopping}
\bibinfo{author}{M.~Safe}, \bibinfo{author}{J.~Carballido},
  \bibinfo{author}{I.~Ponzoni}, \bibinfo{author}{N.~Brignole},
  \bibinfo{title}{On stopping criteria for genetic algorithms}, in:
  \bibinfo{booktitle}{Advances in Artificial Intelligence--SBIA 2004: 17th
  Brazilian Symposium on Artificial Intelligence, Sao Luis, Maranhao, Brazil,
  September 29-Ocotber 1, 2004. Proceedings 17},
  \bibinfo{organization}{Springer}, \bibinfo{pages}{405--413},
  \bibinfo{year}{2004}.

\bibitem[{Jain et~al.(2013)Jain, Bhatele, Robson, Gamblin, and
  Kale}]{10.1145/2503210.2503263}
\bibinfo{author}{N.~Jain}, \bibinfo{author}{A.~Bhatele}, \bibinfo{author}{M.~P.
  Robson}, \bibinfo{author}{T.~Gamblin}, \bibinfo{author}{L.~V. Kale},
  \bibinfo{title}{Predicting Application Performance Using Supervised Learning
  on Communication Features}, in: \bibinfo{booktitle}{Proceedings of the
  International Conference on High Performance Computing, Networking, Storage
  and Analysis}, SC '13, \bibinfo{publisher}{Association for Computing
  Machinery}, \bibinfo{address}{New York, NY, USA}, ISBN
  \bibinfo{isbn}{9781450323789}, \doi{\bibinfo{doi}{10.1145/2503210.2503263}},
  \urlprefix\url{https://doi.org/10.1145/2503210.2503263},
  \bibinfo{year}{2013}.

\end{thebibliography}





\end{document}